# The Firefighter Algorithm: A Hybrid Metaheuristic for Optimization Problems

M.Z. Naser[1,2], A.Z. Naser[3,4]

[1]School of Civil & Environmental Engineering and Earth Sciences (SCEEES), Clemson University, USA
[2]Artificial Intelligence Research Institute for Science and Engineering (AIRISE), Clemson University, USA
E-mail: mznaser@clemson.edu, Website: www.mznaser.com
[3]School of Engineering, University of Guelph, Guelph, ON N1G 2W1, Canada, E-mail: anaser@uoguelph.ca
[4]Department of Mechanical Engineering, University of Manitoba, Winnipeg R3T 5V6, Canada

**Abstract**

This paper presents the Firefighter Optimization (FFO) algorithm as a new hybrid metaheuristic for optimization problems. This algorithm stems inspiration from the collaborative strategies often deployed by firefighters in firefighting activities. To evaluate the performance of FFO, extensive experiments were conducted, wherein the FFO was examined against 13 commonly used optimization algorithms, namely, the Ant Colony Optimization (ACO), Bat Algorithm (BA), Biogeography-Based Optimization (BBO), Flower Pollination Algorithm (FPA), Genetic Algorithm (GA), Grey Wolf Optimizer (GWO), Harmony Search (HS), Particle Swarm Optimization (PSO), Simulated Annealing (SA), Tabu Search (TS), and Whale Optimization Algorithm (WOA), and across 24 benchmark functions of various dimensions and complexities. The results demonstrate that FFO achieves comparative performance and, in some scenarios, outperforms commonly adopted optimization algorithms in terms of the obtained fitness, time taken for exaction, and research space covered per unit of time.

## 1.0 Introduction

Metaheuristics play a large role in the domain of optimization. These algorithms have been renowned for their efficacy in tackling complex and multidimensional problems that can typically be beyond the reach of traditional methods [1]. Metaheuristics are distinguished by their flexibility and robustness, making them particularly suitable for problems where the solution landscape is rugged or poorly defined, such as those expected across diverse disciplines, ranging from engineering and logistics to economics and data science [2]. For example, metaheuristics' versatility enables their engineering application to optimize design parameters for complex systems. In logistics, metaheuristics can help solve scheduling and routing problems. Similarly, metaheuristics can prove beneficial in finance systems to optimize investment portfolios, to name a few.

Metaheuristics can be defined as high-level strategies that coordinate simpler investigative methodologies to explore and exploit the search space efficiently [3]. These strategies are characterized by their reliance on processes that promote some form of balance between exploration of the search space to avoid entrapment in local optima and exploitation mechanisms that refine promising areas to converge toward global optima [4]. Thus, metaheuristics can be thought of as generic and adaptable to a broad spectrum of problems with minimal modification. This can be advantageous to problem-specific algorithms.

Metaheuristics are broadly classified into two categories based on their approach: single-solution based and population-based. Single-solution metaheuristics iteratively improve a single candidate solution. Some such metaheuristics include Simulated Annealing (SA) and Tabu Search (TS). On the other hand, population-based metaheuristics evolve a group (i.e., population) of solutions to leverage interactions within this group to explore and exploit the search space collectively. Some examples that fall under this group include Genetic Algorithms (GA) and Particle Swarm Optimization (PSO).

Similalrly, metaheuristics can also be classified based on their source of inspiration. Some metaheuristics are nature-inspired algorithms, wherein they are inspired by natural phenomena, biological processes, or behaviors observed in animals/plants. These nature-inspired algorithms often mimic survival mechanisms, evolutionary processes, or social behaviors. For example, GA, PSO, and Ant Colony Optimization (ACO) are representative of such classification. Then, some metaheuristics draw inspiration from human behaviors, the physical world, and societal structures. For instance, both TS and SA incorporate actions and behaviors seen in humans (i.e., memory) and physical processes (i.e., annealing in metallurgy) [5].

Despite their versatility and ease of use, metaheuristics can suffer from challenges [6]. One such challenge is drawing a balance between exploration and exploitation capabilities. These capabilities can be crucial for avoiding premature convergence and ensuring the global optimum is reached. Moreover, the stochastic nature of these algorithms often requires multiple runs to achieve consistent results, which can be computationally expensive [7]. Fortunately, the field of metaheuristics continues to grow in response to addressing increasingly complex problems [8]. One notable trend is the hybridization of metaheuristic algorithms, where two or more distinct strategies are combined to exploit their complementary strengths. For instance, hybrid algorithms might combine one algorithm's explorative power with another's intensive exploitation capabilities. Such hybridization can potentially yield solutions that are both diverse and precise.

Still, as computational challenges grow and the need for efficient optimization strategies intensifies, the role of metaheuristics becomes increasingly important and warranted. The ability of metaheuristics to adapt and provide feasible and efficient solutions. Here, we propose a novel algorithm, Firefighter Optimization (FFO), as a hybrid metaheuristic for optimization problems. A number of comparative experiments, supplemented with various metrics, were carried out to validate the effectiveness and competitiveness of FFO. More specifically, the performance of FFO and the other selected algorithms was examined across 540 tests on a wide range of benchmark and test functions. Our experimental results demonstrate the effectiveness and competitiveness of FFO compared to state-of-the-art algorithms.

The rest of the paper is organized as follows: Section 2 presents a description of FFO and explains each component in detail. Sections 3 and 4 describe the aforementioned algorithms and 24 commonly used benchmarking test functions for various complexity levels. Finally, Sections 4 and 5 conclude the paper with a presentation of our comparative results and conclusions/findings learned from our analysis.

## 2.0 Description of the Firefighter Optimization (FFO) algorithm

This section describes the FFO in more detail. We start with a general description and then dive into a more detailed analysis of FFO's functions.

*2.1 General description*

The Firefighter optimization (FFO) algorithm is inspired by the strategies and tactics used by firefighters to combat fires in real-world scenarios. In the face of a fire, firefighters must strategically allocate resources, decide when to focus on extinguishing the flames, and when to protect specific areas or perform rescue operations. More specifically, this optimization technique draws from various aspects of firefighting, including the dynamic allocation of resources, adaptive response to changing conditions, and coordination among multiple firefighters (i.e., agents) to achieve a common goal. Further, the FFO algorithm is designed to balance exploration and exploitation by maintaining diversity within the search space and adapting its parameters to navigate complex optimization landscapes. See Table 1 for a comparison of this algorithm against other commonly used in optimization.

Initialization: The algorithm starts by randomly initializing a population of agents within the specified bounds of the search space. Each agent represents a potential solution to the optimization problem.

Evaluation: Each agent's position is evaluated using the objective function, which measures the fitness or quality of the solution. The best agent, i.e., the one with the lowest fitness value for minimization problems, is identified as the global best solution.

Adaptive Local Search: The FFO algorithm employs an adaptive local search mechanism to refine the positions of the agents. This process involves generating perturbations around the current position of an agent and evaluating the new positions. The perturbations are adjusted adaptively based on the agent's performance and the iteration count. This local search helps agents escape local optima and explore the solution space more effectively.

Crossover and Mutation: To enhance diversity and promote exploration, the FFO algorithm incorporates crossover and mutation operators. The crossover operator allows agents to exchange information, creating new solutions by combining parts of two parent agents. The mutation operator introduces random changes to an agent's position, providing an additional mechanism to explore new areas of the search space.

Perturbation Mechanism: When agents fail to improve over a certain number of iterations, a perturbation mechanism is triggered. This mechanism mimics the adaptive response of firefighters to changing conditions. Agents undergo a larger perturbation, guided by the global best agent, to move towards potentially better regions of the search space. The intensity of the perturbation increases with the number of unsuccessful iterations, allowing the algorithm to escape stagnation and continue searching for optimal solutions.

Adaptive Step Size: The step size used in the local search and perturbation mechanisms is adaptively adjusted based on the algorithm's progress. If the algorithm detects stagnation, the step size is increased to encourage exploration. Conversely, if the algorithm is converging towards a solution, the step size is reduced to fine-tune the search and improve solution accuracy.

Cooling Schedule: The FFO algorithm incorporates a cooling schedule inspired by simulated annealing. The temperature parameter, which controls the acceptance probability of worse solutions during the local search, is gradually reduced over iterations. This allows the algorithm to initially explore more freely and then gradually focus on exploitation as it approaches convergence.

Termination Criteria: The algorithm runs until one or more termination criteria are met. These criteria can include reaching a maximum number of iterations, exceeding a predefined number of iterations without improvement, or achieving a target fitness value. The termination criteria ensure that the algorithm does not run indefinitely and provides a solution within a reasonable time frame.

*2.2 Detailed description*
A more detailed description of FFO's functions is provided herein.

Initialization (__init__)

The initialization function (__init__) of the FirefighterOptimization class sets up the algorithm's parameters, agents, and initial conditions for optimization. These parameters include the objective function, dimensions of the problem, number of agents, maximum iterations, no-improvement limit, and bounds for the search space. Additional parameters stemming from existing algorithms for crossover, mutation, simulated annealing, and perturbation control are also set up. The agents (i.e., solutions) are initialized randomly within the specified bounds, and the best global agent and fitness are identified at the start.

Mathematically, the initialization function involves setting up vectors and matrices that represent the agents in the solution space. The objective function, $f(x)$, is defined over a domain with specified bounds. Here are the key mathematical components:

Objective Function f(x):
- Represents the function to be optimized.
- $f : R_n \rightarrow R$.

Dimension $d$:
- The number of variables in the optimization problem.

Number of Agents N:
- The size of the population, typically denoted as $N$.

Bounds $[l, u]$:
- The lower ($l$) and upper ($u$) bounds for the variables.
- Each agent $x_i$ is initialized as $x_i \sim U(l, u)$.

Crossover and Mutation Probabilities:
- $p_c$ and $p_m$ are the probabilities for crossover and mutation, respectively.

Simulated Annealing Parameters:
- Initial temperature $T_0$ and cooling rate $\alpha$.

**Initialization Process:**
The process can be broken down into the following steps:

<u>Parameter Setup:</u>
```
self.objective_func = objective_func
self.dimension = dimension
self.num_agents = num_agents
self.max_iter = max_iter self.no_improve_limit = no_improve_limit
self.bounds = bounds
```

<u>Agents Initialization:</u> Agents are randomly distributed within the bounds:
```
self.agents = np.random.uniform(bounds[0], bounds[1], (num_agents, dimension))
```

<u>Best Agent Identification:</u> The initial best global agent and its fitness are determined:
```
self.best_global_agent = np.copy(self.agents[np.argmin([self.objective_func(agent) for agent in self.agents])]) self.best_global_fitness = self.objective_func(self.best_global_agent)
```

<u>Other Parameters:</u> Additional parameters like step size, mutation rates, and counters are initialized:
```
self.step_size = step_size
self.crossover_probability = crossover_probability
self.mutation_probability = mutation_probability
self.initial_temp = initial_temp
self.cooling_rate = cooling_rate
self.mutation_rates = np.full(self.num_agents, 0.1)
self.no_improve_counter = 0 self.iteration = 1
```

**Agent Evaluation (**evaluate_agents**)**
The evaluate_agents function assesses the fitness of each agent within the initiated population. This function updates the best global fitness and agent if/when a better solution is found. This function calculates the fitness of each agent based on the objective function and updates the global best agent if an improved solution is identified. This evaluation process also guides the algorithm's search process toward better solutions. Mathematically, the evaluation involves computing the objective function for each agent and identifying the agent with the minimum fitness value.

<u>Objective Function, $f(x)$:</u> The function to be minimized and is evaluated for each agent $x_i$.

<u>Fitness Calculation:</u> For each agent $x_i$, the fitness is $f(x_i)$.

<u>Best Fitness Update:</u> The global best fitness and agent are updated if a new minimum is found.

<u>Evaluation Process</u>
The evaluation process can be broken down into the following steps:

<u>Fitness Calculation:</u> The fitness of each agent is computed:
```
fitness = np.array([self.objective_func(agent) for agent in self.agents])
```

*Best Agent Identification:* The index of the agent with the best fitness is found, and if this fitness is better than the current global best, updates are made:
```
best_index = np.argmin(fitness)
if fitness[best_index] < self.best_global_fitness:
        self.best_global_fitness = fitness[best_index]
        self.best_global_agent = np.copy(self.agents[best_index])
        self.no_improve_counter = 0
else:
        self.no_improve_counter += 1
```
*Return Fitness Values:* The fitness values are returned for further processing:
```
return fitness
```

**Agent Update (update_agents)**

The update_agents function is responsible for evolving the population of firefighters (i.e., agents) to maintain diversity through crossover, mutation, and perturbation operations. This function involves modifying agent positions in the solution space to explore new regions. As inspired by genetic algorithms, this function applies crossover to combine traits from different agents, mutation[1] to introduce random changes, and perturbations to escape local optima. Mathematically, the update process includes the following key components:

*Crossover Operation:* Combines parts of two agents to create new agents.

$$x' = \text{crossover}(x_1, x_2).$$

*Mutation Operation:* Applies random perturbations to agents.

$$x' = x + \epsilon, \text{ where } \epsilon \sim N(0,\sigma).$$

*Perturbation Operation:* Adjusts agents based on the global best agent if no improvement is observed.

$$x' = x + \alpha(x* - x), \text{ where } \alpha \text{ is the perturbation intensity.}$$

*Update Process:* The update process involves the following steps:

*Evaluate Agents*: The fitness of agents is then evaluated to update the global best fitness achieved:
```
self.evaluate_agents()
```

*Crossover and Mutation*: Each agent undergoes crossover and mutation based on set probabilities:
```
for i in range(self.num_agents):
        if np.random.rand() < self.crossover_probability:
                partner_index = np.random.randint(self.num_agents)
                self.agents[i], self.agents[partner_index] = self.crossover(self.agents[i], self.agents[partner_index])
        if np.random.rand() < self.mutation_probability:
                self.agents[i] = self.local_search(self.agents[i], i)
```

*Perturbation*: If no improvement is observed for an extended period, this function applies perturbations:
```
if self.no_improve_counter > 50: self.agents[i] = self.apply_perturbation(self.agents[i], 0.1 + 0.02 * (self.no_improve_counter - 50))
```

---

[1] Further information on these operations will be provided in a subsequent section.

*Boundary Check*: Agents are clipped within the bounds:
`self.agents[i] = np.clip(self.agents[i], self.bounds[0], self.bounds[1])`

*Trajectory and Perturbation History*: Updates to agents are recorded for trajectory analysis:
`self.trajectory.append(np.copy(self.agents[i]))  self.perturbation_history.append(self.agents[i])`

**Local Search (local_search)**

The local_search function refines an agent's position by exploring its neighborhood. This function helps agents escape local optima and find better solutions. This function involves making small adjustments to an agent's position to find a better solution in its vicinity. The process is guided by a temperature parameter, allowing the acceptance of worse solutions early on to escape local optima. As the temperature decreases, the search becomes more focused on local refinement – in a similar process to controlling fires. Mathematically, local search applies perturbations to an agent's position and evaluates the new positions. The acceptance of new positions is probabilistic, influenced by a temperature parameter.

Perturbation:
$x' = x + \epsilon \sim N(0, \sigma)$.

Temperature:
$T = T_0 \cdot \alpha^k$, where $T_0$ is the initial temperature, $\alpha$ is the cooling rate, and $k$ is the iteration.

Acceptance Probability:
$P(\Delta E) = \exp(-\Delta E/T)$, where $\Delta E$ is the change in fitness.

Local Search Process: The process involves the following steps:

*Temperature Calculation:* The temperature is calculated based on the iteration:
`temp = self.initial_temp * (self.cooling_rate ** self.iteration)`

*Local Best Initialization*: The current agent is considered the local best:
`best_local = agent`
`best_local_fitness = self.objective_func(agent)`

*Perturbation and Evaluation:* Small adjustments are made to the agent's position, and new positions are evaluated:
`for _ in range(10 + 5 * (self.no_improve_counter // 100)):`
`    perturbation = np.random.normal(0, self.step_size * self.mutation_rates[index], self.dimension)`
`    candidate = best_local + perturbation`
`    candidate_fitness = self.objective_func(candidate)`

*Acceptance Check:* New positions are accepted based on fitness improvement or probabilistically:
`if candidate_fitness < best_local_fitness or np.random.rand() < np.exp((best_local_fitness - candidate_fitness) / temp):`
`    best_local = candidate`
`    best_local_fitness = candidate_fitness`

*Return Local Best:* The refined local best position is then returned:
`return best_local`

**Perturbation Application (apply_perturbation)**

Perturbation application is a strategy to escape local optima by making larger adjustments to agents' positions, which can be particularly useful when the algorithm stagnates. The apply_perturbation function introduces significant changes to agents' positions based on the global best agent when no improvement is observed. In mathematical notation, perturbation involves adjusting an agent's position towards the global best agent, scaled by an intensity factor, such that:

Direction Vector:
$d = x* - x$, where $x*$ is the global best agent and $x$ is the current agent.

Perturbation:
$x' = x + \alpha \cdot d$, where $\alpha$ is the perturbation intensity.

Perturbation Process
The process involves the following steps:

*Direction Calculation:* The direction vector from the agent to the global best is calculated:
direction = self.best_global_agent - agent

*Perturbation Calculation:* A perturbation is applied based on the direction vector and intensity:
perturbation = np.random.normal(0, intensity, self.dimension) * direction

*New Position Calculation:* The agent's new position is calculated:
return agent + perturbation

**Crossover (**crossover**)**

The crossover function combines parts of two agents to create new agents as a means to introducing diversity into the initialized population. This genetic algorithm-inspired method helps explore new solutions by recombining existing ones. Key components in this function include:

Crossover Point:
A random point $p$ is selected along the agents' dimensions.

New Agents:
New agents are created by combining segments of the parent agents.

Crossover Process: The process involves the following steps:

*Crossover Point Selection:* A random crossover point is selected:
crossover_point = np.random.randint(1, self.dimension)

*Agent Combination:* New agents are created by combining segments of the parent agents:
new_agent1 = np.concatenate((agent1[:crossover_point], agent2[crossover_point:]))
new_agent2 = np.concatenate((agent2[:crossover_point], agent1[crossover_point:]))

*Return New Agents:* The new agents are returned:
return new_agent1, new_agent2

**Cooling Schedule (**cooling_schedule**)**
The cooling_schedule function adjusts the step size based on the algorithm's progress, similar to the *cooling* option in simulated annealing. The cooling schedule involves gradually reducing the step size as the algorithm progresses (in a similar manner to controlling the fire toward the later stages of firefighting). This process allows for a finer search of the solution space over time, balancing exploration and exploitation. Mathematically, the cooling schedule involves updating the step size based on the number of iterations and the no-improvement counter such that:

Step Size Update:
The step size is reduced based on a cooling factor.

Cooling Process:
The process involves the following steps:

*Step Size Adjustment:* The step size is adjusted based on the no-improvement counter:
```
if self.no_improve_counter > 50:
        self.step_size *= 0.98
else:
        self.step_size *= 0.99
```

**Execution Loop (**run**)**

The run function controls the main execution loop of the algorithm, where agents are updated, evaluated, and the cooling schedule is applied until a termination condition is met. It also tracks the best solution and its fitness across iterations. The execution loop is the core of the optimization process. It iteratively updates agents, evaluates their fitness, applies the cooling schedule, and checks termination conditions. This loop continues until the optimization criteria are met, ensuring that the algorithm converges to an optimal solution. Mathematically, the execution loop involves iterating over the update and evaluation processes while tracking the best solution. Key components include:

Iteration:
The algorithm proceeds to iterate up to a maximum number of iterations.

Agent Update and Evaluation:
Agents are updated and evaluated in each iteration.

Cooling Schedule:
The step size is adjusted according to the cooling schedule.

Termination Check:
The algorithm checks whether the termination conditions are met.

Execution Process
The process involves the following steps:

*Initialization*: The fitness history and trajectory are initialized:
```
self.fitness_history = []
self.trajectory = []
self.iteration = 1
self.start_time = time.time()
```

*Main Loop*: The main loop iterates until the termination condition is met:
```
while not self.should_terminate():
   self.update_agents()
```

```
        self.cooling_schedule()
        self.fitness_history.append(self.best_global_fitness)
        if self.verbose:
            print(f"Iteration {self.iteration}, Best Fitness {self.best_global_fitness}, No Improve Counter {self.no_improve_counter}")
        self.iteration += 1
self.end_time = time.time()
```

*Return Best Solution*: The best global agent and fitness are returned:
```
return self.best_global_agent, self.best_global_fitness, self.fitness_history
```

**Termination Check (**should_terminate**)**

The should_terminate function determines whether the algorithm should stop based on several conditions: *maximum iterations reached*, *no improvement for a set number of iterations*, or *achieving a target fitness level*. Thus, this function ensures that the algorithm terminates when further exploration is unlikely to yield better results.

Iteration Check: The algorithm stops if the maximum number of iterations is reached (i.e., $k \geq$ max_iter).

No Improvement Check: The algorithm stops if no improvement is observed for a set number of iterations.
```
no_improve_counter > no_improve_limit
```

Target Fitness Check: The algorithm stops if the target fitness level is achieved.
$f(x*) \leq$ target_fitness

Termination Process
The process involves the following steps:

*Condition Check:* The termination condition is evaluated based on the iteration, no-improvement counter, and best global fitness:
```
termination_condition = (
        self.iteration >= self.max_iter or
        self.no_improve_counter > self.no_improve_limit or
        self.best_global_fitness < self.target_fitness
)
```

*Verbose Output*: If termination is triggered and *verbose*[2] mode is enabled, a message is printed:
```
if termination_condition and self.verbose:
        print(f"Terminating: Iteration={self.iteration}, No Improve Counter={self.no_improve_counter}, Best Fitness={self.best_global_fitness}")
```

*Return Condition*: The termination condition is returned:
```
return termination_condition
```

**Execution Time (**get_execution_time**)**

The get_execution_time function calculates the total runtime of the algorithm as a means to present a measure for evaluating the time efficiency of the algorithm. This execution time is calculated as the difference between the end time and the start time.

---

[2] This option can be turned on/off to include additional prints/debugging.

Start Time: The time when the algorithm begins execution.

End Time: The time when the algorithm completes execution.

Total Execution Time:

```
execution_time = end_time – start_time
```

Execution Time Calculation

*Return Execution Time:* The total execution time is calculated and returned:
```
return self.end_time - self.start_time
```

## Trajectory Tracking (get_trajectory)

The get_trajectory function records the sequence of solutions explored by the algorithm, which can be further analyzed to examine the search behavior and pathway through the solution space. The trajectory involves recording the positions of agents over time.

Trajectory:
A sequence of positions $x_t$ recorded at each iteration.

Position Recording:
Each agent's position is recorded during the update process.

Trajectory Tracking Process
The process involves the following steps:

*Return Trajectory*: The trajectory of agents' positions is returned:
```
return self.trajectory
```

## Total Distance Traveled (get_total_distance)

The get_total_distance function computes the cumulative distance traveled by the algorithm in the solution space. Such a distance can be an indicator of the algorithm's exploratory behavior and efficiency. The total distance traveled involves summing the Euclidean distances between consecutive positions of agents.

Euclidean Distance:
The distance between two points $x_i$ and $x_{i+1}$ is $\|x_{i+1} - x_i\|^2$.

Cumulative Distance:
The total distance is the sum of distances between consecutive positions.

Distance Calculation Process
The process involves the following steps:

*Initialize Distance:* The total distance is initialized:
```
distance = 0
```

*Calculate Distances:* Distances between consecutive positions are calculated and summed:
```
for i in range(1, len(self.trajectory)):
    distance += np.linalg.norm(self.trajectory[i] - self.trajectory[i - 1])
```

*Return Total Distance:* The total distance traveled is returned:
```
return distance
```

```python
class FirefighterOptimization:
    def __init__(self, objective_func, dimension, num_agents=100, max_iter=500, no_improve_limit=30, bounds=(-5.12, 5.12), step_size=1.0, crossover_probability=0.5, mutation_probability=0.1, initial_temp=100.0, cooling_rate=0.95, verbose=False, use_additional_conditions=False, target_fitness=1e-5):
        self.objective_func = objective_func
        self.dimension = dimension
        self.num_agents = num_agents
        self.max_iter = max_iter
        self.no_improve_limit = no_improve_limit
        self.bounds = bounds
        self.agents = np.random.uniform(bounds[0], bounds[1], (num_agents, dimension))
        self.best_global_agent = np.copy(self.agents[np.argmin([self.objective_func(agent) for agent in self.agents])])
        self.best_global_fitness = self.objective_func(self.best_global_agent)
        self.step_size = step_size
        self.crossover_probability = crossover_probability
        self.mutation_probability = mutation_probability
        self.initial_temp = initial_temp
        self.cooling_rate = cooling_rate
        self.mutation_rates = np.full(self.num_agents, 0.1)
        self.no_improve_counter = 0
        self.iteration = 1
        self.fitness_history = []
        self.perturbation_history = []
        self.verbose = verbose
        self.use_additional_conditions = use_additional_conditions
        self.target_fitness = target_fitness
        self.trajectory = []
        self.start_time = None
        self.end_time = None

    def evaluate_agents(self):
        fitness = np.array([self.objective_func(agent) for agent in self.agents])
        best_index = np.argmin(fitness)
        if fitness[best_index] < self.best_global_fitness:
            self.best_global_fitness = fitness[best_index]
            self.best_global_agent = np.copy(self.agents[best_index])
            self.no_improve_counter = 0
        else:
            self.no_improve_counter += 1
        return fitness

    def update_agents(self):
        self.evaluate_agents()
        for i in range(self.num_agents):
            if np.random.rand() < self.crossover_probability:
```

```
            partner_index = np.random.randint(self.num_agents)
            self.agents[i], self.agents[partner_index] = self.crossover(self.agents[i], self.agents[partner_index])
        if np.random.rand() < self.mutation_probability:
            self.agents[i] = self.local_search(self.agents[i], i)
        if self.no_improve_counter > 50:
            self.agents[i] = self.apply_perturbation(self.agents[i], 0.1 + 0.02 * (self.no_improve_counter - 50))
        self.agents[i] = np.clip(self.agents[i], self.bounds[0], self.bounds[1])
        self.trajectory.append(np.copy(self.agents[i]))
        self.perturbation_history.append(self.agents[i])

    def local_search(self, agent, index):
        temp = self.initial_temp * (self.cooling_rate ** self.iteration)
        best_local = agent
        best_local_fitness = self.objective_func(agent)
        for _ in range(10 + 5 * (self.no_improve_counter // 100)):
            perturbation = np.random.normal(0, self.step_size * self.mutation_rates[index], self.dimension)
            candidate = best_local + perturbation
            candidate_fitness = self.objective_func(candidate)
            if candidate_fitness < best_local_fitness or np.random.rand() < np.exp((best_local_fitness - candidate_fitness) / temp):
                best_local = candidate
                best_local_fitness = candidate_fitness
        return best_local

    def apply_perturbation(self, agent, intensity):
        direction = self.best_global_agent - agent
        perturbation = np.random.normal(0, intensity, self.dimension) * direction
        return agent + perturbation

    def crossover(self, agent1, agent2):
        crossover_point = np.random.randint(1, self.dimension)
        new_agent1 = np.concatenate((agent1[:crossover_point], agent2[crossover_point:]))
        new_agent2 = np.concatenate((agent2[:crossover_point], agent1[crossover_point:]))
        return new_agent1, new_agent2

    def cooling_schedule(self):
        if self.no_improve_counter > 50:
            self.step_size *= 0.98
        else:
            self.step_size *= 0.99

    def run(self):
        self.fitness_history = []
        self.trajectory = []
        self.iteration = 1
```

```python
        self.start_time = time.time()
        while not self.should_terminate():
            self.update_agents()
            self.cooling_schedule()
            self.fitness_history.append(self.best_global_fitness)
            if self.verbose:
                print(f"Iteration {self.iteration}, Best Fitness {self.best_global_fitness}, No Improve Counter {self.no_improve_counter}")
            self.iteration += 1
        self.end_time = time.time()
        return self.best_global_agent, self.best_global_fitness, self.fitness_history

    def should_terminate(self):
        if self.use_additional_conditions:
            termination_condition = (
                self.iteration >= self.max_iter or
                self.no_improve_counter > self.no_improve_limit or
                self.best_global_fitness < self.target_fitness
            )
        else:
            termination_condition = self.iteration >= self.max_iter
        if termination_condition and self.verbose:
            print(f"Terminating: Iteration={self.iteration}, No Improve Counter={self.no_improve_counter}, Best Fitness={self.best_global_fitness}")
        return termination_condition

    def get_execution_time(self):
        return self.end_time - self.start_time

    def get_trajectory(self):
        return self.trajectory

    def get_total_distance(self):
        distance = 0
        for i in range(1, len(self.trajectory)):
            distance += np.linalg.norm(self.trajectory[i] - self.trajectory[i - 1])
        return distance
```

Table 1 Qualitative comparison between FFO and commonly used optimization algorithms

| Feature/Aspect | FFO | ACO | BA | BBO | FPA | GA | GWO | HS | PSO | SA | TS | WOA |
|---|---|---|---|---|---|---|---|---|---|---|---|---|
| Inspiration | Firefighting strategies | Ant foraging behavior | Echolocation of bats | Biogeography concepts | Flower pollination | Natural evolution | Grey wolves' hunting | Musical harmony | Swarming behavior of birds/fish | Annealing process in metallurgy | Memory-based search | Whale bubble-net hunting |
| Exploration vs. Exploitation | Balanced through adaptive mechanisms | Strong exploration initially, then exploitation | Balanced | Balanced | Balanced | Exploration initially, then exploitation | Balanced | Balanced | Balanced | Exploration initially, then exploitation | Exploitation-focused | Balanced |
| Memory Usage | Utilizes historical data for perturbation | Pheromone trails as indirect memory | Historical positions of bats | Habitat suitability index | Best solutions pollinate | Population-based, uses historical data | Pack leader memory | Harmony memory | Historical best positions | Simulated states | Tabu list | Whale positions memory |
| Main Operators | Local search, crossover, mutation, perturbation | Pheromone update, path selection | Echolocation, frequency tuning | Migration, mutation | Global and local pollination | Selection, crossover, mutation | Encircling prey, attacking, searching | Pitch adjustment, random selection | Velocity and position update | Temperature-based state changes | Tabu list and neighborhood search | Encircling prey, bubble-net hunting |
| Adaptivity | Adaptive step size and perturbation intensity | Pheromone evaporation rate | Frequency adjustment, loudness and pulse rate | Migration rates, mutation rates | Switching probability | Mutation and crossover probabilities | Adaptation in search phases | Adjustments in harmony memory consideration rate | Inertia weight, cognitive and social coefficients | Temperature and cooling schedule | Adaptive tabu list size | Adaptive hunting mechanism |
| Convergence Speed | Generally fast due to adaptive mechanisms | Moderate | Fast | Moderate | Fast | Fast | Fast | Moderate | Fast | Moderate | Slow to moderate | Fast |
| Computational Complexity | Moderate to high, depends on parameter settings | Moderate | Moderate | Moderate | Moderate | Moderate | Moderate | Moderate | Low to moderate | Low | Moderate to high | Moderate |
| Parameter Sensitivity | Moderately sensitive; requires tuning | Highly sensitive to pheromone parameters | Moderately sensitive | Moderately sensitive | Moderately sensitive | Highly sensitive | Moderately sensitive | Moderately sensitive | Moderately sensitive | Highly sensitive | Moderately sensitive | Moderately sensitive |
| Scalability | Good for high-dimensional problems | Moderate | Good | Good | Good | Good | Good | Good | Good | Moderate | Good | Good |
| Flexibility | High, can incorporate various strategies | Moderate | Moderate | Moderate | Moderate | High | Moderate | Moderate | High | High | Moderate | Moderate |

- Inspiration: The natural or artificial process that inspired the algorithm's development.
- Exploration vs. Exploitation: The algorithm's balance between searching new areas (exploration) and refining known good areas (exploitation).
- Memory Usage: How the algorithm uses past information to guide future searches.
- Main Operators: The primary mechanisms or processes the algorithm uses to find solutions.
- Adaptivity: The algorithm's ability to adjust its parameters dynamically during the optimization process.
- Convergence Speed: How quickly the algorithm typically finds a solution.
- Solution Quality: The effectiveness of the algorithm in finding high-quality solutions.
- Computational Complexity: The computational resources required by the algorithm, often related to time and memory usage.
- Parameter Sensitivity: The degree to which the algorithm's performance is affected by its parameter settings.
- Scalability: The algorithm's capability to handle problems of increasing size or complexity.
- Flexibility: The algorithm's adaptability to different types of optimization problems.
- Diversity Maintenance: How the algorithm ensures a diverse set of solutions to avoid premature convergence.
- Typical Applications: Common fields or problems where the algorithm is frequently applied.

## 3.0 Description of benchmarking algorithms, experiments, and functions

This section describes the experimental examination used to benchmark FFO. For a start, FFO was examined against 13 other commonly used optimization algorithms, namely, the Ant Colony Optimization (ACO), Bat Algorithm (BA), Biogeography-Based Optimization (BBO), Cuckoo Search (CS), Firefly Algorithm (FA), Flower Pollination Algorithm (FPA), Genetic Algorithm (GA), Grey Wolf Optimizer (GWO), Harmony Search (HS), Particle Swarm Optimization (PSO), Simulated Annealing (SA), Tabu Search (TS), and Whale Optimization Algorithm (WOA). All of these algorithms were used in their default settings[3], and a brief description of each is presented herein for completion. We invite interested readers to review the original publications for these algorithms to learn more about their settings and applications.

### *3.1 Ant Colony Optimization (ACO)*

The Ant Colony Optimization (ACO) was formulated by Dorigo and colleagues [9] based on the pheromone-laying behavior observed in certain ant species. This method uses ants as artificial agents to simulate the decision-making process of real ants in selecting paths. As ants traverse paths, they deposit pheromones that guide subsequent ants toward promising solutions. The ACO algorithm is a probabilistic approach to problem-solving where the search space is represented as a graph, and paths through this graph are evaluated based on the intensity of pheromone deposits. The algorithm's efficiency hinges on several parameters: alpha (influence of pheromone on path selection, set at 1.0), beta (influence of heuristic information on path choice, set at 2.0), and rho (rate of pheromone evaporation, set at 0.5). This algorithms has been successfully applied to network routing, scheduling, and other optimization problems that involve finding optimal paths through graphs [10].

### *3.2 Bat Algorithm (BA)*

Developed by Yang et al. in 2012 [11], the Bat Algorithm (BA) was designed to mimic the echolocation behavior of bats. This algorithm models bats that emit sound waves to navigate and locate prey, translating this biological mechanism into a search and optimization strategy. In BA, each simulated bat adjusts its flight based on velocity, loudness, and echolocation frequency, which dynamically changes from exploration to exploitation phases depending on the proximity to optimal solutions. Key parameters of BA include alpha (initial loudness, set at 0.5), gamma (rate of loudness decrease and emission rate increase, set at 0.5), and frequency range ($f_{min}$ at 0, $f_{max}$ at 2.0). BA is adept at tackling complex problems characterized by continuous and multimodal search spaces and has shown effectiveness in engineering design and dynamic optimization tasks [12].

### *3.3 Biogeography-Based Optimization (BBO)*

Introduced by Simon in 2008 [13], Biogeography-Based Optimization (BBO) leverages migration concepts from biogeography to solve optimization problems. BBO operates on the premise that species migrate between habitats, affecting their survival and reproduction rates. The algorithm features migration operators that simulate gene flow by exchanging solution features, akin to species migration in nature. Key components include the habitat suitability index, which evaluates the desirability of solutions, and migration rates that determine the exchange intensity between solutions. BBO also uses mutation to enhance genetic diversity and avoid premature convergence on suboptimal solutions. The BBO algorithm has proven effective in network design, power systems optimization, and other applications where geographic considerations are crucial [14].

---

[3] The default setting for FFO include:
FirefighterOptimization:
    def __init__(self, objective_func, dimension, num_agents=100, max_iter=500, no_improve_limit=30, bounds=(-5.12, 5.12), step_size=1.0, crossover_probability=0.5, mutation_probability=0.1, initial_temp=100.0, cooling_rate=0.95, verbose=False, use_additional_conditions=False, target_fitness=1e-5):)

### 3.4 Cuckoo Search (CS)
Cuckoo Search (CS) was developed by Yang and Deb [15]. This algorithm is inspired by the obligate brood parasitism of some cuckoo species by laying their eggs in the nests of other host birds. If a host bird discovers the eggs are not its own, it will either throw them away or abandon its nest. The algorithm uses this idea to lay a new solution (egg) into a randomly chosen nest, and the best nests with high-quality eggs will be carried over to the next generations. CS is known for its simplicity and flexibility. It has been effectively applied in solving problems like structural design, scheduling, and routing problems where the search space is discrete, and the global optimum is hidden among many local optima [16].

### 3.5 Firefly Algorithm (FA)
The flashing behavior of fireflies inspires the Firefly Algorithm (FA). Such a flashing behavior acts as a signal system to attract other fireflies. FA, developed by Yang in 2008 [17], uses these biologically inspired techniques to handle optimization problems and functions. Fireflies in the algorithm search the space by moving towards brighter and more attractive fireflies. The attractiveness is proportional to the brightness, and both decrease as their distance increases. The landscape of the objective function determines the brightness of a firefly. A key advantage of FA is its ability to deal with multimodal optimization problems, as it naturally divides the population into subgroups that converge to different optima. Some of the key settings in the FA algorithm include alpha (a randomness factor that affects the movement of a firefly and helps fireflies explore the search space beyond the immediate neighboring fireflies, selected at 0.5), beta (controls how strongly other fireflies are drawn towards it, selected at 1.0), and gamma (influences how the attractiveness of a firefly decreases with distance, selected at 1.0). This feature makes it particularly useful for complex functions with multiple local optima. FA has been applied to problems like economic dispatch, clustering, and image processing [18].

### 3.6 Flower Pollination Algorithm (FPA)
Devised by Yang et al. in 2012 [19], the Flower Pollination Algorithm (FPA) algorithm emulates the natural pollination processes of flowers. It aims to optimize solutions by alternating between self-pollination and cross-pollination mechanisms that are naturally facilitated by natural vectors like insects, wind, or water. This approach maintains solution diversity and promotes effective convergence. In FPA, solutions are represented as flowers whose attractiveness—determined by fitness—guides the pollination process. The algorithm uses local pollination for minor adjustments within the immediate search area and global pollination, employing Levy flights for broader searches to escape local optima. FPA's dual strategy has been effectively applied to engineering design and economic load dispatch challenges [20].

*3.7 Genetic Algorithm (GA)*

Holland [21] formulated the Genetic Algorithm (GA) as a computational analog to natural selection, embodying the principle of survival of the fittest. GA begins with a population of randomly generated individuals, evolving over generations to optimize solutions. Selection is fitness-based, favoring solutions that perform better under a defined fitness function. GA incorporates mutation and crossover as genetic operators to introduce variability and new traits into offspring. Typical parameters include a population size of 100, a mutation rate of 0.1, and a crossover rate of 0.1. This algorithm is widely utilized across fields such as optimization, automatic programming, and machine learning, where it helps solve complex problems efficiently [22].

*3.8 Grey Wolf Optimizer (GWO)*

The Grey Wolf Optimizer (GWO), introduced by Mirjalili et al. in 2014 [23], is a nature-inspired metaheuristic algorithm inspired by grey wolves' social structure and hunting behavior. Grey wolves exhibit a distinct hierarchical system consisting of alpha, beta, delta, and omega wolves, with each tier playing a specific role within the pack. In GWO, this hierarchy is mirrored in the solution process: the alpha wolf represents the optimal solution, followed by beta and delta as the second and third best solutions, respectively, while omega wolves embody the remaining candidate solutions. The algorithm leverages this structure to simulate the wolves' hunting strategy, which is segmented into three phases: tracking, encircling, and attacking prey, each reflecting a critical phase of the optimization process. GWO is adept at navigating complex, multidimensional landscapes, making it valuable in fields such as mechanical engineering design and renewable energy optimization, where the search spaces often exhibit high nonlinearity and multimodality.

*3.9 Harmony Search (HS)*

Developed by Geem et al. in 2001 [24], Harmony Search (HS) is an optimization algorithm inspired by the improvisational process of musicians tuning their instruments to achieve aesthetic harmony. This algorithm iteratively adjusts solution vectors in a similar fashion to musicians' adjust pitches to optimize a given function. HS employs a stochastic approach rather than a gradient-based method, enhancing its efficacy in addressing non-differential and discrete problems. The algorithm's performance is governed by two primary parameters: the harmony memory consideration rate (set at 0.9), which dictates the likelihood of selecting existing memory solutions for new harmonies, and the pitch adjustment rate (set at 0.3), which determines how much the chosen solutions are modified. HS has shown significant utility in solving complex engineering problems such as structural and water network design, where traditional methods may struggle due to the extensive search spaces involved.

*3.10 Particle Swarm Optimization (PSO)*

Particle Swarm Optimization (PSO) was introduced by Kennedy and Eberhart in 1995 [25]. This algorithm simulates the social behaviors observed in flocks of birds or schools of fish. This metaheuristic optimizes problem solutions by iteratively enhancing a population of candidate solutions based on the personal and collective experiences of the particles. For example, the PSO starts with randomly initialized particles (solutions) and updates their positions within the search space by balancing personal best achievements and global knowledge shared across the swarm.

The algorithm is known for its simplicity and adaptability, often requiring few parameter adjustments. It utilizes three key parameters: swarm size (typically 100 particles), cognitive coefficient (influence of the particle's own memory, set at 1.0), and social coefficient (influence of neighboring particles, also set at 1.0). These factors influence the dynamics of particle movements towards optimal solutions. PSO is particularly effective in continuous, high-dimensional environments and has been applied successfully across various domains, including electrical power systems, robotics, and bioinformatics [26].

*3.11 Simulated Annealing (SA)*
Simulated Annealing (SA) was developed by Kirkpatrick et al. in 1983 and Cerny in 1985 [27]. This is a probabilistic method designed to approximate the global optimum of a function. SA is inspired by the metallurgical process of annealing, where materials are heated and then gradually cooled to improve their structural properties. This process is mimicked by allowing a system to explore higher energy states (solutions) by heating, thereby overcoming local optima, followed by slow cooling to stabilize at a lower energy state (optimal solution). The algorithm makes random transitions to neighboring solutions, accepting improvements outright and worse solutions based on a decreasing probability over time. Key parameters include the initial temperature (set at 100) and cooling rate (set at 0.95). SA has been successfully applied across various domains, from economics to computational science [28].

*3.12 Tabu Search (TS)*
Tabu Search (TS) was introduced by Glover in the late 1980s [29] as a metaheuristic that extends beyond local search methods. TS extends such methods by adopting a memory structure, the tabu list, to avoid cycling back to previously encountered suboptimal solutions. This list temporarily bans certain moves, helping the algorithm to escape local optima and explore less favorable solutions that might lead to a globally optimal solution. TS is adaptable, allowing periodic resetting of the tabu status to balance exploration and exploitation. It is particularly effective in complex scheduling, logistical planning, and assignment problems, where its ability to navigate challenging solution spaces is applications [30].

*3.13 Whale Optimization Algorithm (WOA)*
The Whale Optimization Algorithm (WOA) was developed by Mirjalili and Lewis in 2016 [31]. The WOA draws inspiration from the bubble-net feeding behavior of humpback whales. This algorithm simulates the whales' strategies of encircling prey and using a spiral path to close in, which are mirrored in the shrinking encircling mechanism and spiral updating position phases of the algorithm. These methods allow the WOA to balance exploration and exploitation dynamically, making it adept at handling complex, non-separable, and nonlinear optimization problems. WOA's effectiveness is demonstrated in its applications across mechanical design and industrial engineering, where it optimizes a variety of challenging problem landscapes [32].

**4.0 Description of utilized benchmarking functions**
This section describes 24 benchmark functions commonly used in benchmarking analysis.

*4.1 Ackley Function*
The Ackley function has a two-dimensional form with a relatively uniform plane [33]. This function also has several dozen local minimums and one global extreme of significantly smaller value than most of the local minimums. This function allows very efficient testing of optimization algorithms as regards stopping at local extremes. The Ackley function is designed to test the ability of optimization algorithms to escape local minima and converge towards a global minimum in a complex landscape. The global minimum is at $x=0$ where $f(x)=0$. This function has the following form:

$$f(x) = -20 \exp\left(-0.2\sqrt{\frac{1}{n}\sum_{i=1}^{n} x_i^2}\right) - \exp\left(\frac{1}{n}\sum_{i=1}^{n} \cos(2\pi x_i)\right) + 20 + e \qquad \text{Eq. 1}$$

*4.2 Alpine Function*
The Alpine function is a multimodal and non-smooth function and hence can provide significant challenges in terms of local minima and ruggedness tests [34]. This function examines the ability of optimization algorithms to handle non-differentiable points with abrupt changes. The global minimum for this function occurs at $x=0$ where $f(x)=0$. This function can be useful for testing optimization algorithms in real-world problems involving non-smooth dynamics, such as mechanical systems with friction or other resistive forces. This function has the following form:

$$f(x) = \sum_{i=1}^{n} |x_i \sin(x_i) + 0.1 x_i| \qquad \text{Eq. 2}$$

*4.3 Booth's Function*
This function presents a simple test case for algorithm testing with a convex with a single global minimum at $(x,y) = (1,3)$ where $f(x,y) = 0$. This function has the following form:

$$f(x, y) = (x + 2y - 7)^2 + (2x + y - 5)^2 \qquad \text{Eq. 3}$$

*4.4 Cross-in-Tray Function*
This function is known for its challenging landscape, characterized by a high degree of multimodality [35]. The function contains several deep holes, indicative of global minima, which are located symmetrically in the function's domain, and may present a significant challenge in the convergence process of algorithms to navigate complex landscapes and avoid local minima in favor of locating and confirming global minima. The global minima occur at approximately $(x,y)=(1.34941,−1.34941),(−1.34941,1.34941),(1.34941,1.34941),(−1.34941,−1.34941)$ with $f(x,y)≈−2.06261$. This function has the following form:

$$f(x, y) = -0.0001 \left(\left|\sin(x)\sin(y)\exp\left(\left|100 - \frac{\sqrt{x^2+y^2}}{\pi}\right|\right)\right| + 1\right)^{0.1} \qquad \text{Eq. 4}$$

*4.5 Drop-Wave Function*
This function features a rippled wave surface that turn challenging to optimize due to its frequent local minima and a pronounced global minimum [36]. This is a multimodal function with a global minimum at $(x,y)=(0,0)$ where $f(x,y)=−1$. It tests an algorithm's capability to navigate through frequent oscillations to find the lowest point and is particularly relevant for simulations and optimizations in fields involving vibrational analysis and wave propagation (i.e., acoustics and materials science, etc.). This function has the following form:

$$f(x, y) = -\frac{1+\cos(12\sqrt{x^2+y^2})}{0.5(x^2+y^2)+2}$$ Eq. 5

### 4.6 Easom Function

The Easom function is a highly unimodal benchmark function stemming from its narrow global peak that is surrounded by a flat landscape [37]. This function has a constant plane over the vast majority of the domain with one global minimum, at $(x,y)=(\pi,\pi)$ where $f(x,y)=-1$, that is difficult to locate due to the flatness of the surrounding area. Oftentimes, this function is used in testing the precision and convergence characteristics of optimization algorithms, and their ability to hone in on and precisely converge to a sharply defined minima. This function has the following form:

$$f(x, y) = -\cos(x)\cos(y)\exp(-((x-\pi)^2 + (y-\pi)^2))$$ Eq. 6

### 4.7 Eggholder Function

The Eggholder function has a highly irregular and complex surface characterized by an uneven plane with several local minimums (of values like its only global minimum) [38,39]. The global minimum is found at $(x,y)=(512,404.2319)$ where $f(x,y)\approx-959.6407$. This function is frequently used in benchmarking sophisticated global optimization algorithms, especially those intended to solve rugged and unpredictable landscape-based problems. This function has the following form:

$$f(x, y) = -(y + 47)\sin\left(\sqrt{\left|\left(\frac{y+x}{2+47}\right)\right|}\right) - x\sin(\sqrt{|x - (y + 47)|})$$ Eq.7

### 4.8 Expanded Schaffer's F6 Function

This is an expansion of the original Schaffer's function that hopes to test for algorithm effectiveness over a broader area with a more complex landscape. More specifically, the function is highly multimodal and oscillatory and hence presents a significant challenge in identifying the global minimum amidst numerous local minima. The function has a global minimum at $(x,y)=(0,0)$ where $f(x,y)=0$. This function can be used in testing spatial algorithms that may be applied in fields like geographic information systems and molecular dynamics where spatial relationships and dynamics are crucial.

$$f(x) = 0.5 + \frac{\sin^2(\sqrt{x^2+y^2})-0.5}{[1+0.001(x^2+y^2)]^2}$$ Eq. 8

### 4.9 Expanded Zakharov Function

This is an extension of the Zakharov function [34] and provides a more challenging scenario for testing optimization algorithms by combining linear, quadratic, and quartic terms. This function has a single global minimum at $x=0$ where $f(x)=0$. This function is used to evaluate the performance of large-scale optimization algorithms in areas such as financial modeling and energy systems, where complex interactions between variables are common.

$$f(x) = \sum_{i=1}^{n} x_i^2 + (\sum_{i=1}^{n} 50ix_i)^2 + (\sum_{i=1}^{n} 50ix_i)^4$$ Eq. 9

*4.10 Goldstein-Price Function*
This function was created by Goldstein and Price [40] to provide a multimodal complex landscape with sharp peaks and valleys. The same function has several local minima and a global minimum found at $(x,y) = (0,-1)$ with $f(x,y) = 3$. This function has the following form:

$$f(x,y) = [1 + (x + y + 1)^2(19 - 14x + 3x^2 - 14y + 6xy + 3y^2)] \times [30 + 2(sx - 3y)^2(18 - 32x + 12x^2 + 48y - 36xy + 27y^2)] \quad \text{Eq. 10}$$

*4.11 Griewank Function*
The Griewank function has large, flat areas interrupted by periodic narrow, deep valleys [41]. This function is highly multimodal and oscillatory and can challenge the algorithm's ability to find the global minimum (at $x=0$) amidst frequent changes in gradient. It is particularly used to test the efficiency of algorithms in handling complex oscillations and multimodal functions with application within acoustic waveguides design and structural engineering with regard to vibrations. This function has the following form:

$$f(x) = 1 + \sum_{i=1}^{n} \frac{x_i^2}{4000} - \prod_{i=1}^{n} \cos(x_i/\sqrt{i}) \quad \text{Eq. 11}$$

*4.12 Himmelblau's Function*
Developed by Himmelblau [42], this function is characterized by multiple global minima, which makes it interesting for testing the robustness of optimization algorithms to locate and distinguish between multiple optima within a complex landscape. This function has four identical global minima located at $(x,y)=(3,2),(-2.805118,3.131312),(-3.779310,-3.283186),(3.584428,-1.848126)$ where $f(x,y)=0$. This function has the following form:

$$f(x,y) = (x^2 + y - 11)^2 + (x + y^2 - 7)^2 \quad \text{Eq. 12}$$

*4.13 Holder Table Function*
The Holder Table function features several deep and narrow global minima and is designed to challenge optimization algorithms in finding and recognizing global solutions in a multimodal space [43]. The function's global minima are symmetrically located around the origin, with four known global minima where $f(x,y)=-19.2085$. This function can serve as a good evaluation metric for multimodal optimization capabilities in logistics and routing problems where multiple equivalent optimal routes need to be evaluated. This function has the following form:

$$f(x,y) = -\left|\sin(x)\cos(y)\exp\left(\left|1 - \frac{\sqrt{x^2+y^2}}{\pi}\right|\right)\right| \quad \text{Eq. 13}$$

*4.14 Levy Function N.13*
This is a variant to the Levi function. This variant is designed to test algorithms against steep gradients and local optima. More specifically, this function has steep ridges and a complex global structure with a global minimum at $(x,y)=(1,1)$ where $f(x,y)=0$. The Levi can be useful in examining algorithms that need to handle sudden changes in gradient effectively. This function has the following form:

$$f(x,y) = \sin^2(3\pi x) + (x-1)^2(1+\sin^2(3\pi y)) + (y-1)^2(1+\sin^2(2\pi y)) \qquad \text{Eq. 14}$$

*4.15 Matyas Function*

This function was created by Matyas [44] as convex function that could serve as an elementary test case for basic functionality and efficiency of optimization algorithms in a controlled setting. The Matyas function has a global minimum at $(x,y) = (0,0)$ where $f(x,y)=0$. This function has the following form:

$$f(x,y) = 0.26(x^2 + y^2) - 0.48xy \qquad \text{Eq. 15}$$

*4.16 Michalewicz Function*

This function was created by Michalewicz [45] to be especially difficult for evolutionary algorithms to solve. This function is highly multimodal, with sharp peaks and valleys that are sensitive to the variable $m$, which controls the steepness of the valleys and ridges. The function's global minimum becomes more difficult to locate as $m$ increases (with a typical value of $m=10$). This function can be used in the testing and development of genetic and evolutionary algorithms, particularly effective for applications requiring high precision in aerodynamics and biomechanical engineering. This function has the following form:

$$f(x) = -\sum_{i=1}^{n} \sin(x_i) \sin^{2m}\left(\frac{i x_i^2}{\pi}\right) \qquad \text{Eq. 16}$$

*4.17 Rastrigin Function*

This function is named after Rastrigin [46] and presents an example of a highly non-linear multimodal function with frequent local minima. The global minimum is at $x=0$ where $f(x)=0$. The function is particularly designed to test the algorithm's capability to escape local minima and is widely used in the testing and development of algorithms in evolutionary computation and real-world scenarios where noise and local minima are prevalent, such as in electronic circuit design. This function has the following form:

$$f(x) = 10n + \sum_{i=1}^{n}[x_i^2 - 10\cos(2\pi x_i)] \qquad \text{Eq. 17}$$

*4.18 Rosenbrock Function*

This function was designed by Rosenbrock in 1960 as a non-linear and non-convex function to test the performance of optimization algorithms over rugged terrain with a narrow, curved valley leading to a global minimum [47]. This function has a global minimum inside a long, narrow, parabolic shaped flat valley. In general, finding the valley is straightforward but converging to the global minimum is difficult. A common multidimensional generalization of this function has the following form:

$$f(x) = \sum_{i=1}^{n-1}[100(x_{i+1} - x_i^2)^2 + (1 - x_i)^2] \qquad \text{Eq. 18}$$

*4.19 Schaffer Function N. 2*

This function is a variant belonging to the family of functions introduced by Schaffer, which are used to evaluate the performance of optimization algorithms in handling oscillating landscapes with narrow valleys [48]. The function is non-convex and multimodal, with a global minimum at $(x,y)$=(0,0) where $f(x,y)$=0. This variant can be sensitive to initial conditions due to the presence of sharp peaks and deep valleys. This function has the following form:

$$f(x,y) = 0.5 + \frac{\sin^2(x^2-y^2)-0.5}{[1+0.001(x^2+y^2)]^2} \qquad \text{Eq. 19}$$

*4.20 Schwefel Function*
The Schwefel function is a classic optimization test problem introduced by Schwefel [49]. This function is initially created for evolutionary algorithms and has complex and non-linear large number of local minima (with a global minimum located near the bounds of the search space at $x$=(420.9687,…,420.9687) where $f(x)$≈0). This function can be helpful in in fields such as aerospace for optimizing the shapes and trajectories of dynamic flying bodies. This function has the following form:

$$f(x) = 418.9829n - x_i \sin(\sqrt{|x_i|}) \qquad \text{Eq. 20}$$

*4.21 Sphere Function*
The Sphere function is one of the classical and simplest benchmark functions often used to test preliminary optimization algorithms in terms of convergence and accuracy. The Sphere function is continuous, convex, and unimodal with a global minimum at $x^* = 0$ where $f(x^*) = 0$. This function has the following form:

$$f(x) = \sum_{i=1}^{n} x_i^2 \qquad \text{Eq. 21}$$

*4.22 Styblinski-Tang Function*
The Styblinski-Tang function has steep valleys and multiple local minima and hence is recognized for its utility in testing optimization algorithms [50]. This function is multimodal, with each variable contributing quadratically and quartically to the output. This function can be suitable for evaluating the efficiency of algorithms in high-dimensional spaces and their capability to scale with increasing dimensionality (which makes it appropriate for practical engineering problems involving material design and circuit optimization with multiple variables required to be simultaneously optimized). The global minimum for this function is at $x_i$=−2.903534$x$ (for all $i$) where $f(x)$=−39.16599$n$ (where $n$ is the number of dimensions). This function has the following form:

$$f(x) = \frac{1}{2}\sum_{i=1}^{n}(x_i^4 - 16x_i^2 + x_i 5) \qquad \text{Eq. 22}$$

*4.23 Three-hump Camel Function*
This function is named for its shape, resembling three camel humps and is designed as a simple and effective benchmark for testing optimization algorithms [51]. This function can be used to evaluate an algorithm's capability to escape local minima and find the global minimum. The Three-hump Camel function has a global minimum at $(x,y)$=(0,0) where $f(x,y)$=0. This function has the following form:

$$f(x,y) = 2x^2 - 1.05x^4 + \frac{x^6}{6} + xy + y^2 \qquad \text{Eq. 23}$$

*4.24 Whitley's Function*

Whitley's function is known for its complex landscape with a high number of local minima [39]. This function is tests algorithms for their ability to distinguish subtle gradient changes and avoid premature convergence. This function can be used for complex, real-world problems such as landscape exploration and molecular configuration. This function has the following form:

$$f(x) = \sum_{i=1}^{n} \sum_{j=1}^{n} \left(\frac{100(x_i^2 - x_j)^2 + (1-x_j)^2}{4000} - \cos\left(200(x_i^2 - x_j)^2 + (1-x_j)^2\right) + 1\right) \quad \text{Eq. 24}$$

Table 2 includes the name of each function, its typical domain range, primary characteristics, known challenges, and the dimensionality for which the function is typically evaluated. Figure 1 compares these functions visually.

Table 2 Holistic comparison of the examined functions.

| Function Name | Domain Range | Characteristics | Challenges | Typical Dimensionality | Minima/ Minimum |
|---|---|---|---|---|---|
| Ackley | (-5, 5) | Nearly flat outer region, large hole | Global minimum difficult to find due to flatness | Multiple | 0 |
| Alpine | (-10, 10) | Multimodal, peaks and valleys | Peaks make locating global minima challenging | Multiple | 0 |
| Beale's | (-4.5, 4.5) | Multimodal | Several local minima | 2 | 0 |
| Booth's | (-10, 10) | Smooth, few local minima | Simplicity, limited challenge | 2 | 0 |
| Cross-in-Tray | (-10, 10) | Highly multimodal | Multiple global minima spread out | 2 | −2.0626 |
| Drop-Wave | (-5.12, 5.12) | Central peak surrounded by ring of minima | Central peak difficult to stabilize on | Multiple | -1 |
| Easom | (-100, 100) | Narrow peak, unimodal | Extremely narrow attraction region | 2 | -1 |
| Eggholder | (-512, 512) | Large search space, numerous local minima | Complex landscape, many minima | Multiple | -959.640 |
| Expanded Schaffer's F6 | (-10, 10) | Numerous local minima, complex landscape | Maintaining algorithm stability | 2 | 0 |
| Expanded Zakharov | (-10, 10) | Combines parabolic and linear terms | Optimization of combined effects | Multiple | 0 |
| Goldstein-Price | (-2, 2) | Complex topology, multimodal | Local minima near boundaries | 2 | 3 |
| Griewank | (-600, 600) | Many widespread minima | Large search space, many local minima | Multiple | 0 |
| Himmelblau's | (-5, 5) | Multiple global minima | Identifying correct global minimum | 2 | 0 |
| Holder Table | (-10, 10) | Symmetric, multiple global minima | Symmetry can confuse algorithms | Multiple | -19.2085 |
| Levy N.13 | (-10, 10) | Steep valleys, complex structure | Pronounced ridges and steep drops | 2 | 0 |
| Matyas | (-10, 10) | Smooth, unimodal | Limited complexity | 2 | 0 |
| Michalewicz | (0, π) | Designed for evolutionary algorithms | Sharp, narrow valleys | Multiple | Multiple |
| Rastrigin | (-5.12, 5.12) | Highly multimodal, oscillating | Numerous local minima, large search space | Multiple | 0 |
| Rosenbrock | (-2.048, 2.048) | Non-convex, narrow curved valley | Finding the global minimum in the narrow valley | Multiple | 0 |
| Salomon's | (-100, 100) | Strong global structure, multimodal | Multiple deceptive local minima | Multiple | 0 |
| Schaffer N. 2 | (-100, 100) | Sharp ridges, flat areas | Balancing exploration and exploitation | Multiple | 0 |
| Schwefel | (-500, 500) | Sinusoidal, maxima and minima | Identifying global minimum amid deceptive maxima | Multiple | 0 |
| Sphere | (-5.12, 5.12) | Smooth, convex, unimodal | Simplicity may not challenge advanced algorithms | Multiple | 0 |
| Styblinski-Tang | (-5, 5) | Steep valleys, multimodal | Harsh penalties for incorrect solutions | Multiple | $-39.165n$ |
| Three-hump Camel | (-5, 5) | Three humps, unimodal | Misleading local minima | 2 | 0 |
| Whitley's | (-10, 10) | Highly complex and multimodal | Complexity and size of search space | Multiple | 0 |

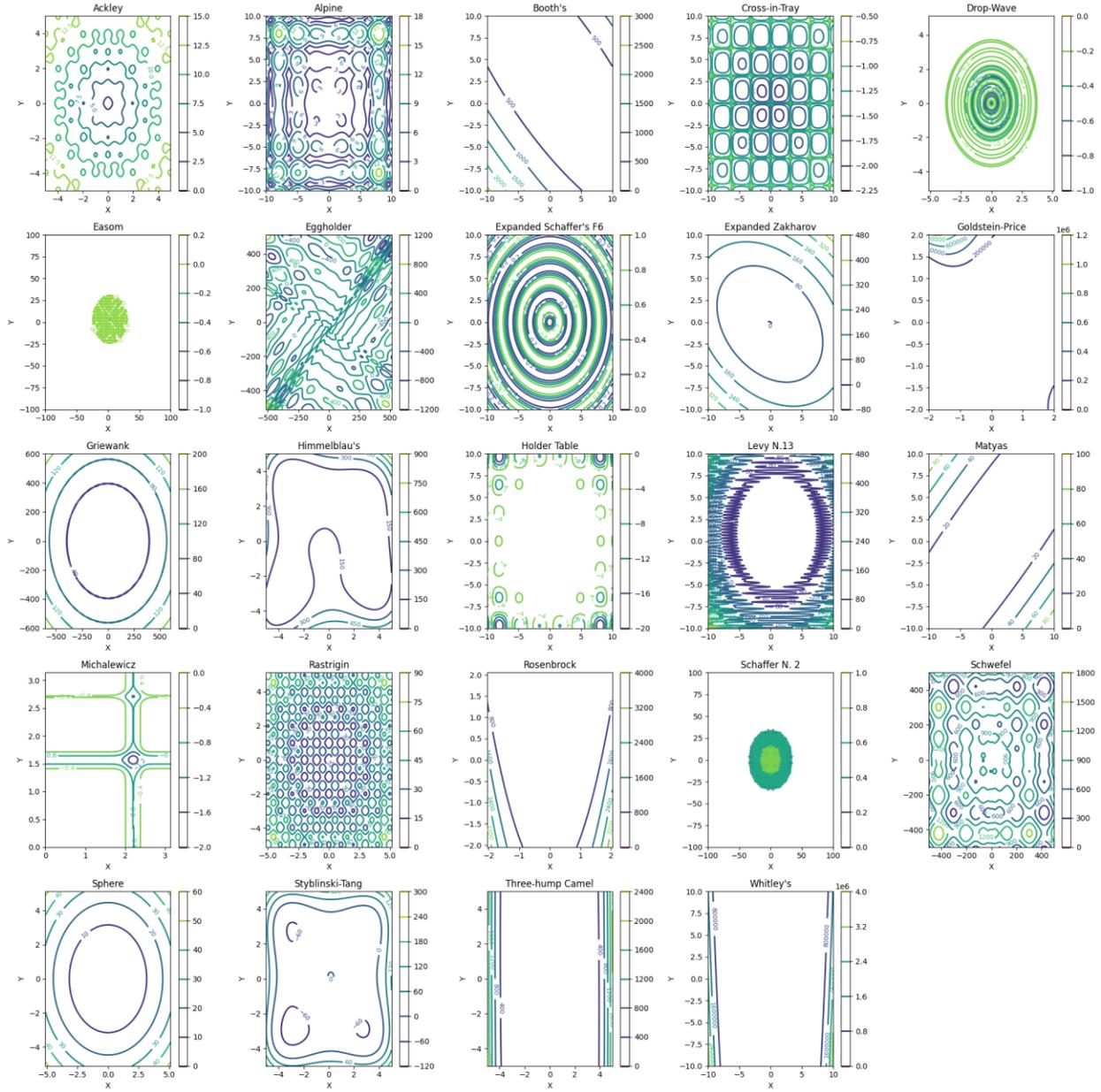

Fig. 1 Visual representation of the utilized test and benchmark functions

## 5.0 Experiments, results, and analysis

Two sets of experiments were conducted to evaluate the performance of the FFO algorithm. In the first, we examine the performance of the FFO algorithm and the other listed algorithms against the aforenoted benchmark functions in 2D settings. Then, we examine the algorithmic performance against the scalable functions at higher dimensions (20D and 50D). In all cases, the experiments entitled a comparison of algorithmic performance in terms of best fitness and fitness history achieved, time taken to execute the analysis per algorithm, trajectory analysis, and a combined combination of metrics. All of these items are discussed herein in detail.

The best fitness is defined as the most favorable value of the objective function obtained in an analysis by an algorithm across all of its iterations or agents. This metric is traced progressively, with the algorithm updating this metric whenever an improved solution is found. Then, the exaction time is defined as the total time from the start of its execution until it terminates. Trajectory refers to the sequence of points that document the position of agents (or the best agent) in the search space. This distance is calculated by summing up the Euclidean distances between consecutive points in the trajectory. Further, four additional categories were used to evaluate the selected algorithms. These include documenting the performance of algorithms in terms of identifying the functions that took the longest (and shortest) time to solve and those that achieved the most (and least) accuracy.

To facilitate a leaner comparison, the execution time and distance explored metrics were combined into a new metric named the Distance per Unit Time metric. This metric directly measures the average distance covered per unit of time, which is directly interpretable as:

$$Distance\ per\ Unit\ Time\ metric = \frac{Total\ distance}{Excution\ time}$$

This measures how much distance is covered on average per second. Therefore, higher values indicate more efficient exploration of the search space.

The conducted analysis was ran and evaluated in a Python 3.10.5 environment using an Intel(R) Core(TM) i7-9700F CPU @ 3.00GHz and an installed RAM of 32.0GB. To ensure fairness, the control parameters of FFO and the 13 metaheuristics employed in the performance evaluation simulation were presented earlier and are found in our simulation script. In all cases, all algorithms ran for 100, 1000, and 3000 iterations and with 10, 50, and 100 agents. As mentioned above, the first leg of the analysis focused on all benchmarking functions at 2D, and the second leg focused on scalable functions (12 out of the original 24 functions) at higher dimensions (20D and 50D).

*2D setting*

Table 3 and Fig. 2 list the overall obtained results from the analysis carried out on all algorithms and functions in 2D setting. This table presents a comparative performance analysis of various optimization algorithms based on metrics such as best fitness, execution time, and distance metrics. Ant Colony Optimization, for instance, shows a mean best fitness of 1.22E+05 with high variability (standard deviation of 9.48E+05) and ranges from -563 to 7.35 in its best fitness performance. In terms of spatial efficiency, it maintains a low average distance metric (mean of 0.092). On the other hand, the Firefly algorithm demonstrates a broader range in execution time, peaking at 1803 seconds, and exhibits a significantly wider spread in the distance metric, reaching up to 91,600. The FFO with additional conditions[4] "OFF" significantly improves best fitness and execution efficiency and large exploration capabilities to a maximum of 1.20E+08. It is quite clear that the FFO ranked well in all metrics and achieved the top ranking in terms of the Distance per Unit Time metric.

---

[4] The FFO (with additional conditions OFF) runs the FFO in a similar fashion to the other algorithms (i.e., to the same number of iterations without any additional stopping criteria). The counterpart version (with additional conditions ON) runs the FFO with all stopping criteria as described above. A true comparison between all algorithms should rely on the OFF version and hence is maintained herein.

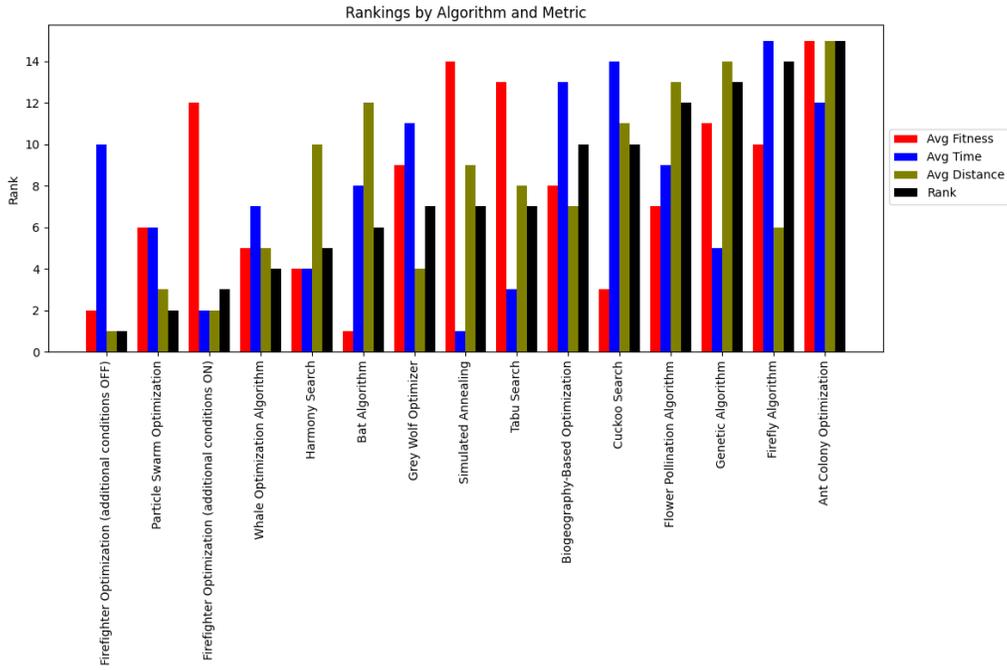

Fig. 2 Ranking of algorithms in 2D settings

Table 3 Overall results for 2D setting

| Algorithm | Best Fitness | | | | Execution Time (s) | | | | Distance metric | | | | Distance per Unit Time |
|---|---|---|---|---|---|---|---|---|---|---|---|---|---|
| | mean | std | min | max | mean | std | min | max | mean | std | min | max | |
| Ant Colony Optimization | 1.22E+05 | 9.48E+05 | -5.63E+02 | 7.35E+00 | 7.35E+00 | 9.39E+00 | 9.00E-02 | 3.51E+01 | 9.26E-02 | 1.36E+00 | 0.00E+00 | 2.00E+01 | 1.26E-02 |
| Bat Algorithm | -1.83E+00 | 4.58E+02 | -9.77E+02 | 2.63E+00 | 2.63E+00 | 3.67E+00 | 1.88E-02 | 1.99E+01 | 3.15E+01 | 1.58E+02 | 0.00E+00 | 1.14E+03 | 1.20E+01 |
| Biogeography-Based Optimization | -9.72E+00 | 1.93E+02 | -8.81E+02 | 7.34E+00 | 7.34E+00 | 9.76E+00 | 7.13E-02 | 4.42E+01 | 2.64E+03 | 1.17E+04 | 0.00E+00 | 1.41E+05 | 3.60E+02 |
| Cuckoo Search | -4.46E+01 | 1.92E+02 | -9.60E+02 | 5.99E+01 | 5.99E+01 | 1.18E+02 | 5.19E-02 | 7.51E+02 | 8.48E+01 | 3.83E+02 | 0.00E+00 | 3.74E+03 | 1.42E+00 |
| FFO (additional conditions OFF) | -4.03E+01 | 1.84E+02 | -9.60E+02 | 4.61E+00 | 4.61E+00 | 7.82E+00 | 2.50E-02 | 5.23E+01 | 2.64E+06 | 1.21E+07 | 1.47E+03 | 1.20E+08 | 5.74E+05 |
| FFO (additional conditions ON) | -2.04E+01 | 1.20E+02 | -8.18E+02 | 9.11E-02 | 9.11E-02 | 1.25E-01 | 0.00E+00 | 7.27E-01 | 1.02E+05 | 3.86E+05 | 0.00E+00 | 3.36E+06 | 1.12E+06 |
| Firefly Algorithm | -2.25E+01 | 1.71E+02 | -8.75E+02 | 1.73E+02 | 1.73E+02 | 3.09E+02 | 1.31E-01 | 1.80E+03 | 7.31E+02 | 6.83E+03 | 0.00E+00 | 9.16E+04 | 4.23E+00 |
| Flower Pollination Algorithm | -3.85E+01 | 1.75E+02 | -9.41E+02 | 2.86E+00 | 2.86E+00 | 3.93E+00 | 2.33E-02 | 2.07E+01 | 6.87E+01 | 3.08E+02 | 0.00E+00 | 2.39E+03 | 2.40E+01 |
| Genetic Algorithm | -3.02E+01 | 1.65E+02 | -9.56E+02 | 1.92E+00 | 1.92E+00 | 2.48E+00 | 2.40E-02 | 1.17E+01 | 5.62E+00 | 1.44E+01 | 3.62E-03 | 1.14E+02 | 2.93E+00 |
| Grey Wolf Optimizer | -1.51E+01 | 1.78E+02 | -9.60E+02 | 5.38E+00 | 5.38E+00 | 7.49E+00 | 4.14E-02 | 4.01E+01 | 3.59E+03 | 1.62E+04 | 8.90E-02 | 1.44E+05 | 6.67E+02 |
| Harmony Search | -3.95E+01 | 1.78E+02 | -9.60E+02 | 7.52E-01 | 7.52E-01 | 1.19E+00 | 6.00E-03 | 7.33E+00 | 1.18E+02 | 4.52E+02 | 0.00E+00 | 3.42E+03 | 1.57E+02 |
| Particle Swarm Optimization | -3.76E+01 | 1.74E+02 | -9.60E+02 | 2.39E+00 | 2.39E+00 | 3.13E+00 | 2.46E-02 | 1.43E+01 | 1.07E+05 | 9.96E+05 | 3.74E+01 | 1.44E+07 | 4.45E+04 |
| Simulated Annealing | -3.80E+00 | 1.34E+02 | -7.18E+02 | 3.65E-02 | 3.65E-02 | 3.41E-02 | 9.97E-04 | 1.25E-01 | 1.04E+02 | 9.39E+01 | 2.30E+00 | 7.66E+02 | 2.86E+03 |
| Tabu Search | 1.02E+01 | 1.81E+02 | -7.87E+02 | 8.23E-02 | 8.23E-02 | 7.41E-02 | 3.99E-03 | 2.63E-01 | 5.77E+02 | 5.14E+02 | 2.89E+01 | 1.51E+03 | 7.01E+03 |
| Whale Optimization Algorithm | -4.14E+01 | 1.78E+02 | -9.58E+02 | 2.45E+00 | 2.45E+00 | 3.22E+00 | 2.47E-02 | 1.47E+01 | 3.73E+03 | 1.30E+04 | 2.94E-02 | 9.68E+04 | 1.52E+03 |

Now, we gain further insights into the performance of all algorithms across different settings through the illustration presented in Fig. 3. This figure shows a sample of trends collected on the selected functions to show how the algorithms perform. It is quite clear that the FFO performs similarly to other algorithms. For example, the first sub-figure highlights performance variations of several optimization algorithms on the Ackley function at 2D, with a specific focus on the number of agents employed. The FFO with additional conditions "OFF" consistently maintains lower fitness values across all agent configurations compared to most algorithms, suggesting higher efficiency. This is in contrast to algorithms like Bat algorithm and Ant Colony Optimization, which show a marked increase in average best fitness. The FFO with additional conditions "OFF" outperforms the latter at higher agent counts, which indicates better scalability with increasing agents. Similar observations can also be made in terms of the algorithmic performance with the number of iterations – especially in the case of the Egg Holder function, which shows superior performance for the FFO algorithm.

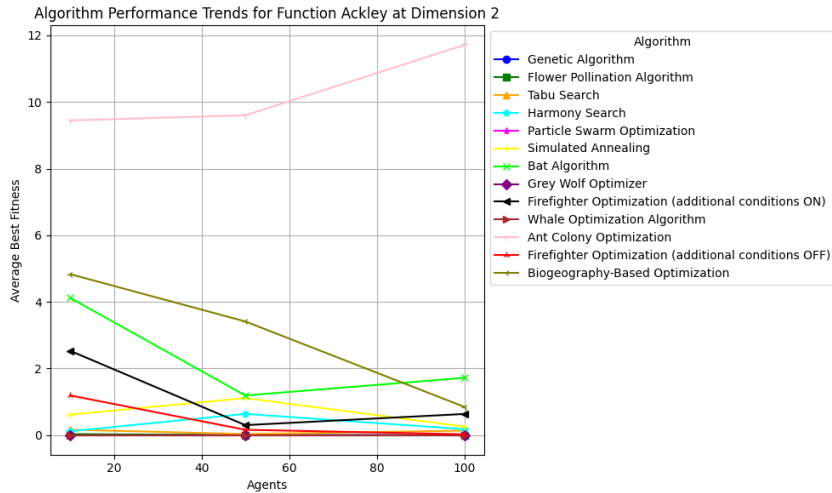

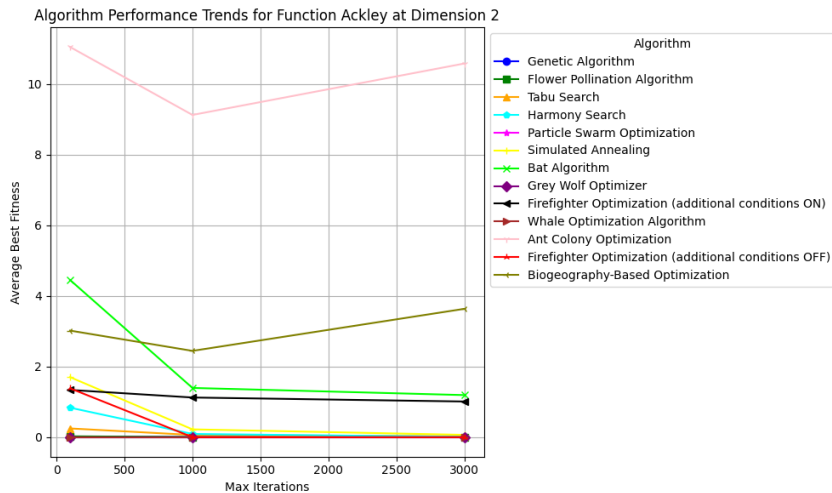

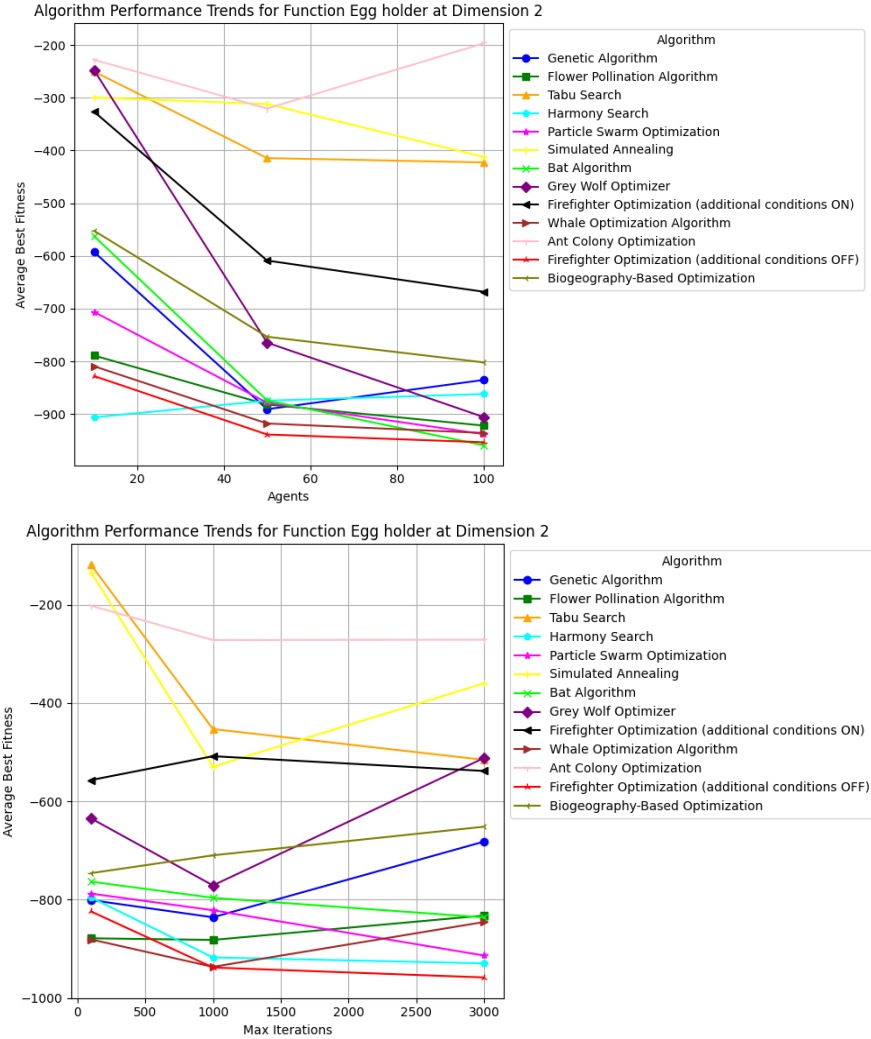

Fig. 3 Sample of trends across different experimental settings

Here, we analyze and rank the performance of optimization algorithms across various settings, specifically focusing on the dimensions, maximum iterations, and the number of agents involved in terms of identifying those that took the longest (and shortest) time to solve, and that achieved the most (and least) accuracy. This process was conducted at each unique setting. First, we rank the algorithms by returning the top three entries of the specified metric and setting. Then, we aggregate these frequencies both locally (for each setting) and globally (across all settings) to provide a comprehensive view of which algorithms consistently perform well or underperform across varied configurations.

We report that the Firefly algorithm appears predominantly in the longest to solve category with a total of 27 out of 27 occurrences. This suggests that the Firefly algorithm, despite its potential advantages in exploring complex landscapes, tends to have longer execution times compared to other algorithms in the study. This could be due to its inherent characteristics, such as the attractiveness parameter and light intensity, which might cause slower convergence, especially in scenarios involving complicated objective functions. The FFO (additional conditions ON) algorithm consistently appears as the fastest solver, also with 27 instances. This algorithm was followed by the Tabu Search. The success in this category indicates the algorithmic potential for applications requiring quick solutions where computational resources can be limited.

In the most accurate category, the Cuckoo Search leads with 9 occurrences, followed by the FFO (additional conditions OFF) with 5 occurnaces and various other algorithms. The notable performance of Cuckoo Search could be attributed to its unique search capabilities, leveraging Lévy flights for global search combined with a probabilistic switch to local search. On the other hand, the Ant Colony Optimization is predominantly featured in the least accurate category with 25 instances. This might indicate ACO's challenges in maintaining high accuracy across the settings tested, potentially due to its reliance on pheromone trails, which might lead to premature convergence or difficulty in escaping local optima in certain types of problems.

Other sets of comparisons between selected algorithms can be seen in Fig. 4. For example, Fig. 4a compares the average history fitness across all functions and for all algorithms. This figure shows that the FFO (additional conditions OFF) demonstrates a rapid convergence initially compared to others like the Bat and Grey Wolf Optimizer, which exhibit more gradual improvements. Figure 4b compares the best fitness achieved in FFO, Tabu Search, the Bat, and the Grey Wolf Optimizer algorithms. In this figure, FFO (additional conditions OFF) excels in deeply multi-modal landscapes like the Ackley and Griewank functions. This suggests that FFO (additional conditions OFF) is particularly adept at managing and escaping local optima in complex search spaces. The graph also shows minimal variance in performance across different configurations (agent counts and iteration limits), indicating the robustness and effectiveness of this algorithm.

Figure 4c shows the performance of all algorithms in tackling continuous and non-continuous functions to provide a clear comparison of how each algorithm handles different types of function landscapes. Here, the FFO (additional conditions OFF) shows comparable performance in both categories, underscoring its versatility. More specifically, in continuous functions, it ranks among the top performers, closely competing with algorithms like Particle Swarm Optimization and Genetic Algorithm. The FFO (additional conditions OFF) also shows similar performance in non-continuous functions. These figures further show the comparative performance and consistency of the newly proposed FFO algorithm against those well established methods.

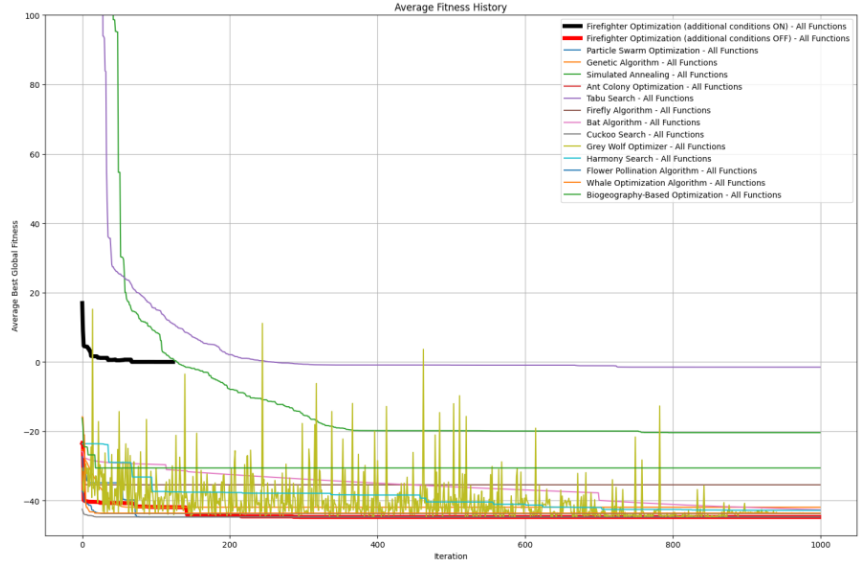

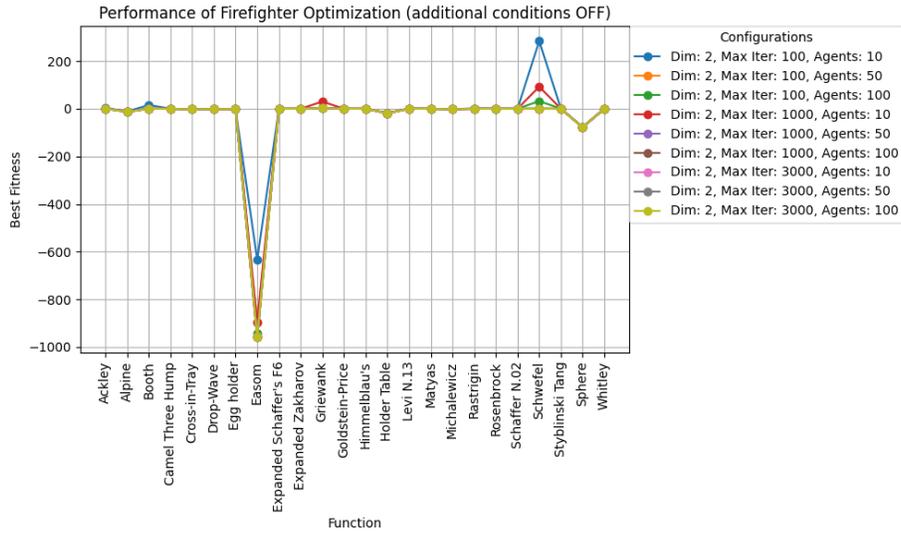

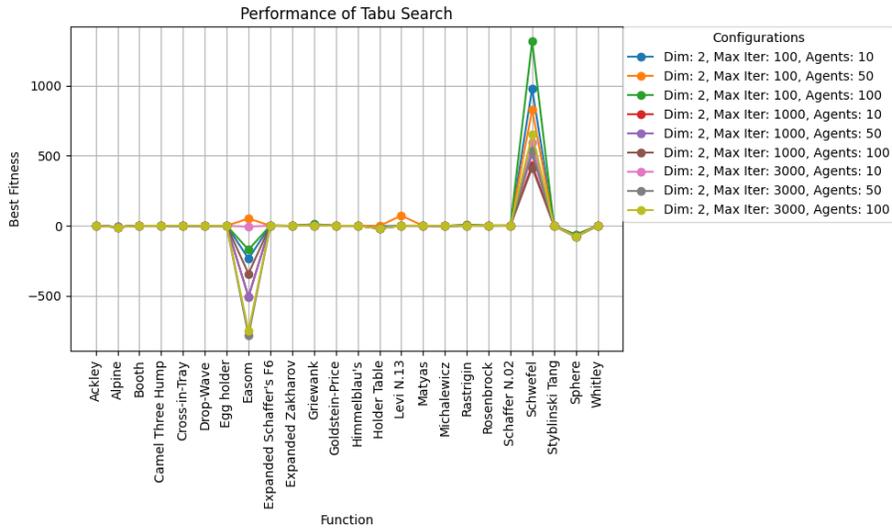

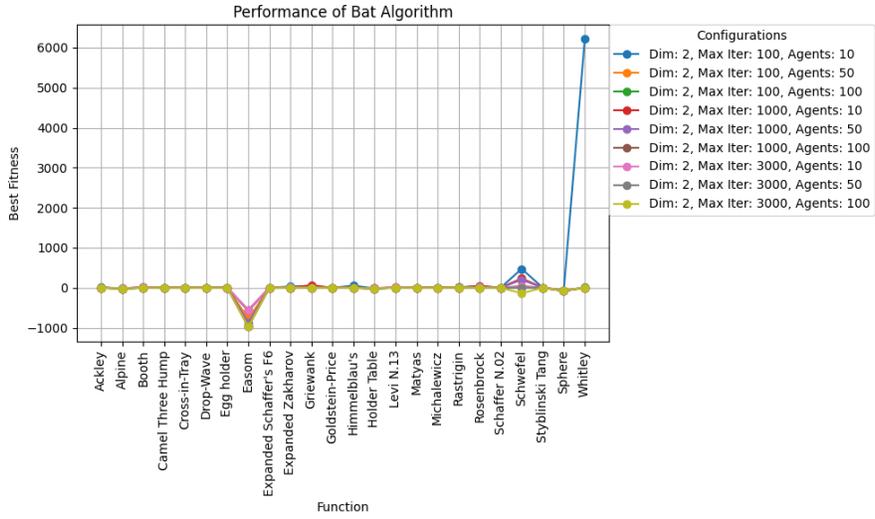

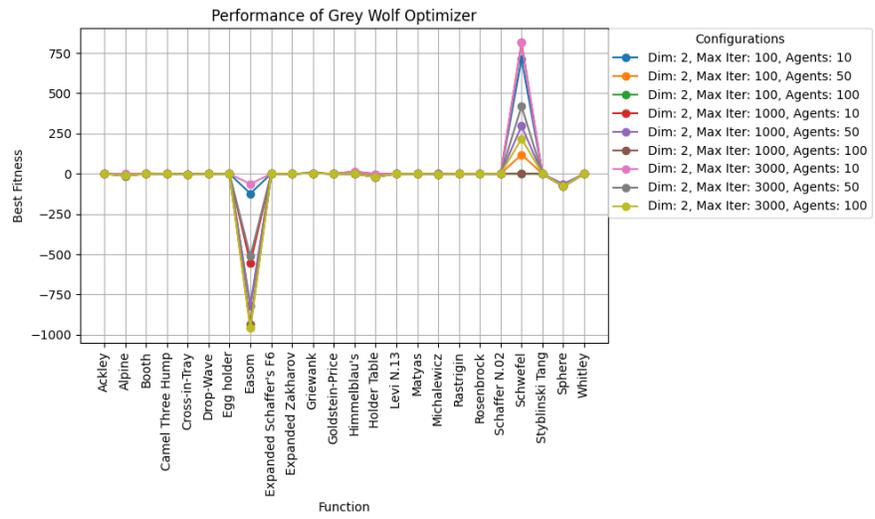

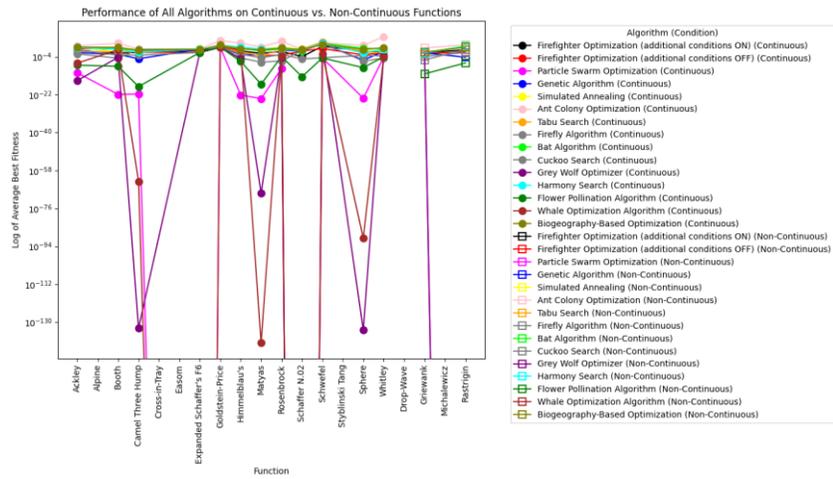

Fig. 4 Further cross examination between the FFO and other notable algorithms

*20D and 50D setting*

Similar to the previous analysis and discussion, Table 4 and Fig. 5 list the overall obtained results from the analysis carried out on all algorithms and functions in higher dimensions of 20D and 50D. Thus, only scalable functions were used in this examination. These functions include Ackley, Eggholder, Easom, Expanded Shaffer's F6, Expanded Zakharov, Griewank, Gldstien-Price, Rosenbrock, Schaffer N.02, Schwefel, Sphere, and Whitely.

In this comparative analysis, the FFO (additional conditions OFF) demonstrates a creditable mean best fitness of 2.88E+03, which, while not the lowest in our dataset, offers a viable trade-off between fitness achievement and computational resources when compared to algorithms like the Simulated Annealing, which has a lower mean best fitness of 5.17E-02 but at a significantly reduced complexity. In terms of execution time, the FFO (additional conditions OFF) records a mean of 9.35E+02 seconds, positioning it as a middle-range performer. It is notably faster than high-accuracy contenders such as the Firefly algorithm and the Biogeography-Based Optimization, which clocks in at mean times of 2.93E+02 seconds and 1.06E+01 seconds, respectively, reflecting a more efficient performance considering the relatively lower fitness figures.

On a more positive note, the FFO (additional conditions OFF) has signifncat exploration capability, as measured by the Total Distance metric, where it posts a mean of 1.58E+07. This is significantly higher than that of the Particle Swarm Optimization and Grey Wolf Optimizer, which stand at 2.98E+05 and 3.10E+04, respectively. Moreover, the Distance per Unit Time for the FFO (with conditions OFF) is large and stands at 2.95E+06, shadowing those of Cuckoo Search and Harmony Search, which report 3.41E+01 and 2.85E+03, respectively. This metric highlights the FFO's efficiency in covering large distances in the search space per unit of time, reinforcing its utility in expansive and complex problem spaces where speed and breadth of exploration are paramount. It is quite clear that the FFO ranked well in all metrics (see Fig. 6).

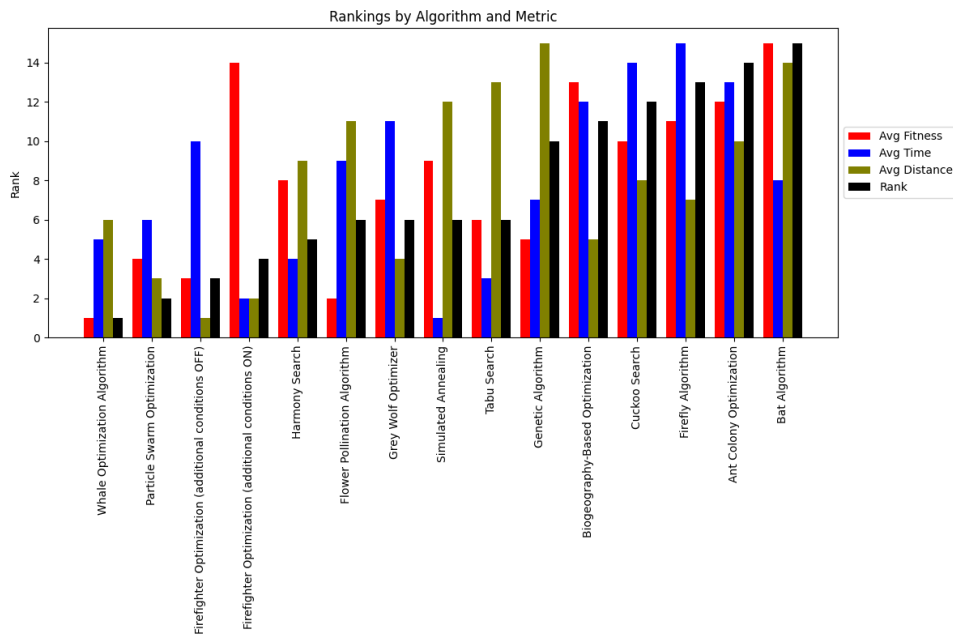

Fig. 6 Ranking of algorithms in 20D and 50D settings

Table 4 Overall results for 20D and 50D setting

| Algorithm | Best Fitness | | | | Execution Time (s) | | | | Distance metric | | | | Distance per Unit Time |
|---|---|---|---|---|---|---|---|---|---|---|---|---|---|
| | mean | std | min | max | mean | std | min | max | mean | std | min | max | |
| Ant Colony Optimization | 7.61E+01 | 1.41E+02 | 9.04E-02 | 6.78E+02 | 9.42E+05 | 1.02E+07 | -9.60E+02 | 1.48E+08 | 1.59E+03 | 4.57E+03 | 0.00E+00 | 3.20E+04 | 2.09E+01 |
| Bat Algorithm | 4.07E+00 | 8.43E+00 | 1.80E-02 | 7.49E+01 | 1.58E+04 | 7.26E+04 | -9.77E+02 | 5.13E+05 | 9.75E+01 | 7.05E+02 | 0.00E+00 | 9.91E+03 | 2.39E+01 |
| Biogeography-Based Optimization | 1.06E+01 | 1.74E+01 | 7.30E-02 | 1.37E+02 | 1.22E+04 | 5.86E+04 | -9.45E+02 | 4.44E+05 | 2.69E+04 | 9.81E+04 | 0.00E+00 | 1.24E+06 | 2.55E+03 |
| Cuckoo Search | 1.20E+02 | 3.66E+02 | 5.37E-02 | 3.63E+03 | 4.22E+03 | 2.62E+04 | -9.60E+02 | 2.99E+05 | 4.10E+03 | 1.23E+04 | 0.00E+00 | 9.31E+04 | 3.41E+01 |
| FFO (additional conditions OFF) | 5.35E+00 | 1.00E+01 | 2.62E-02 | 8.16E+01 | 2.88E+03 | 2.19E+04 | -9.60E+02 | 2.70E+05 | 1.58E+07 | 6.46E+07 | 1.50E+03 | 6.49E+08 | 2.95E+06 |
| FFO (additional conditions ON) | 1.81E-01 | 2.90E-01 | 0.00E+00 | 2.67E+00 | 9.01E+03 | 4.40E+04 | -9.02E+02 | 3.41E+05 | 5.80E+05 | 1.95E+06 | 0.00E+00 | 1.78E+07 | 3.21E+06 |
| Firefly Algorithm | 2.93E+02 | 7.74E+02 | 1.36E-01 | 7.35E+03 | 1.03E+04 | 5.17E+04 | -9.48E+02 | 5.16E+05 | 2.06E+03 | 9.80E+03 | 0.00E+00 | 1.02E+05 | 7.03E+00 |
| Flower Pollination Algorithm | 4.43E+00 | 8.81E+00 | 2.39E-02 | 7.74E+01 | 4.27E+02 | 2.82E+03 | -9.60E+02 | 4.03E+04 | 8.12E+02 | 2.66E+03 | 3.89E-02 | 2.16E+04 | 1.83E+02 |
| Genetic Algorithm | 3.50E+00 | 5.87E+00 | 2.41E-02 | 4.73E+01 | 4.01E+02 | 2.17E+03 | -9.60E+02 | 1.86E+04 | 9.72E+01 | 2.11E+02 | 0.00E+00 | 1.95E+03 | 2.78E+01 |
| Grey Wolf Optimizer | 8.40E+00 | 1.72E+01 | 4.20E-02 | 1.52E+02 | 7.23E+02 | 3.46E+03 | -9.60E+02 | 2.09E+04 | 3.10E+04 | 1.05E+05 | 6.21E-02 | 8.51E+05 | 3.68E+03 |
| Harmony Search | 1.83E+00 | 4.06E+00 | 6.99E-03 | 3.69E+01 | 4.15E+03 | 2.89E+04 | -9.60E+02 | 3.48E+05 | 5.22E+03 | 2.10E+04 | 0.00E+00 | 2.12E+05 | 1.57E+02 |
| Particle Swarm Optimization | 3.19E+00 | 5.31E+00 | 2.50E-02 | 4.31E+01 | 4.80E+02 | 2.67E+03 | -9.60E+02 | 3.35E+04 | 2.98E+05 | 9.34E+05 | 3.24E+01 | 1.05E+07 | 4.45E+04 |
| Simulated Annealing | 5.17E-02 | 7.16E-02 | 9.97E-04 | 5.57E-01 | 4.58E+03 | 3.05E+04 | -8.21E+02 | 3.73E+05 | 5.18E+02 | 7.21E+02 | 1.78E+00 | 4.96E+03 | 2.86E+03 |
| Tabu Search | 1.40E+00 | 3.07E+00 | 3.91E-03 | 2.14E+01 | 7.04E+02 | 3.04E+03 | -7.53E+02 | 2.18E+04 | 5.04E+02 | 5.25E+02 | 2.17E+01 | 2.04E+03 | 7.01E+03 |
| Whale Optimization Algorithm | 3.19E+00 | 5.25E+00 | 2.54E-02 | 4.18E+01 | -5.35E+01 | 4.24E+02 | -9.60E+02 | 6.14E+03 | 2.10E+04 | 6.20E+04 | 9.85E-02 | 5.91E+05 | 1.52E+03 |

Now, we gain further insights into the performance of all algorithms across the settings of 20D and 50D, as seen in Fig. 7. It can be seen that the FFO performs consistently similarly to other algorithms at different settings of agents and iterations.

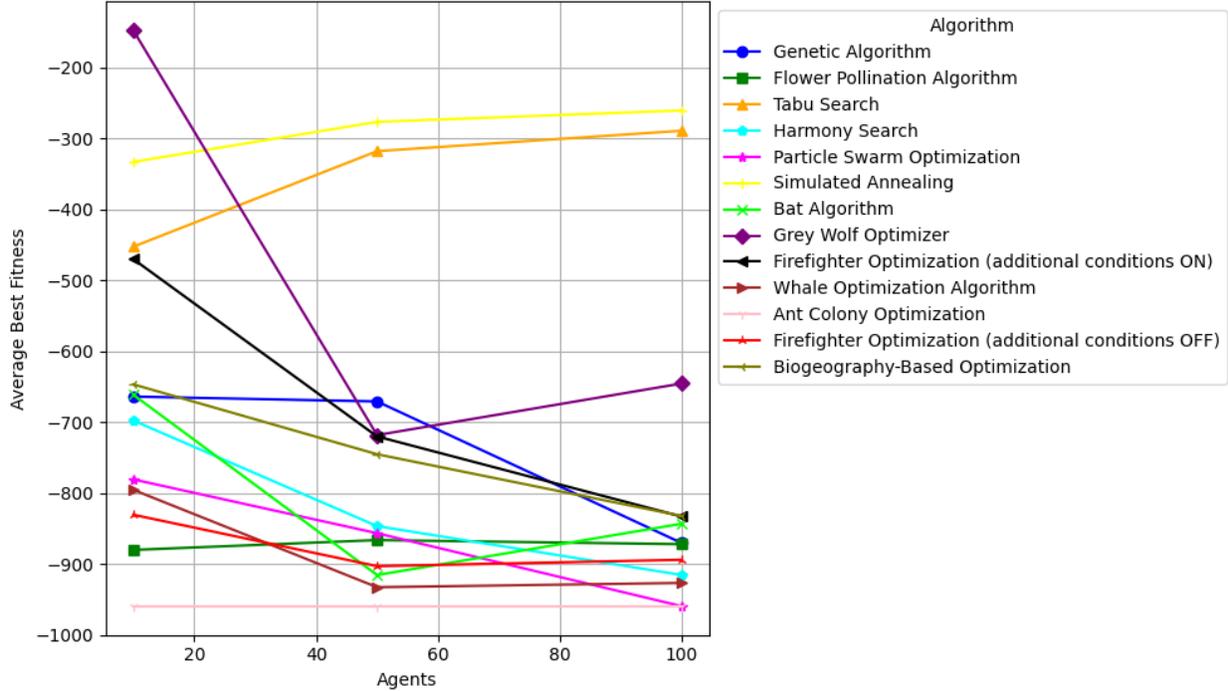

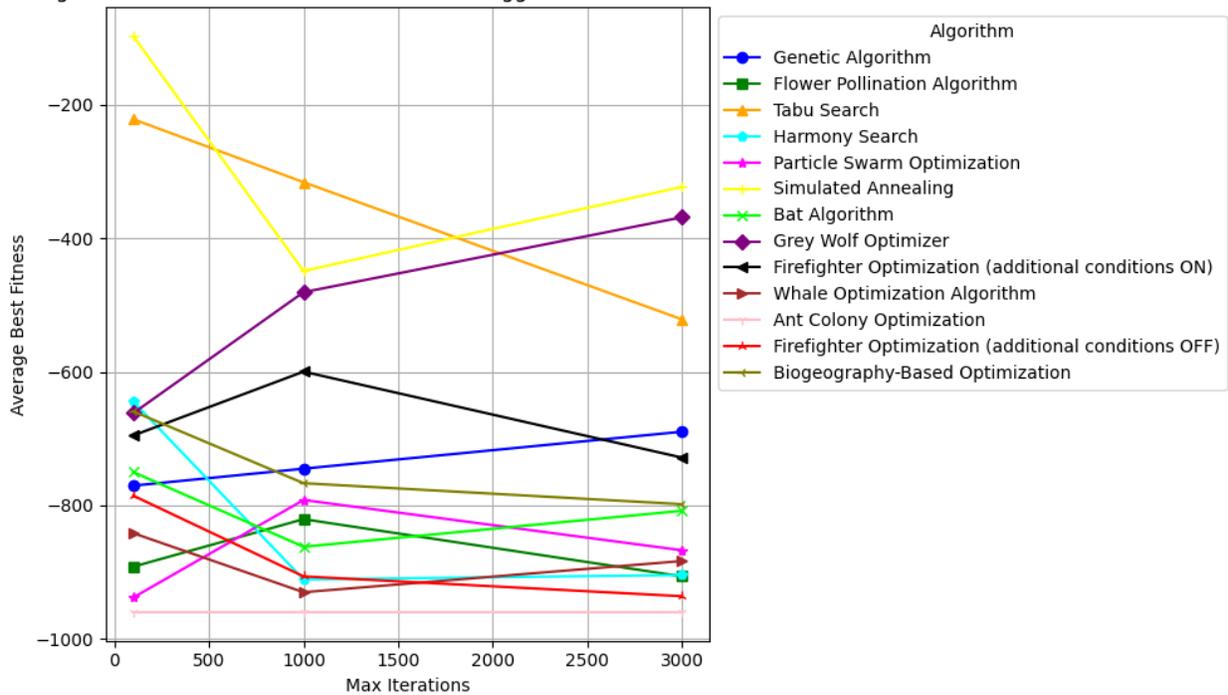

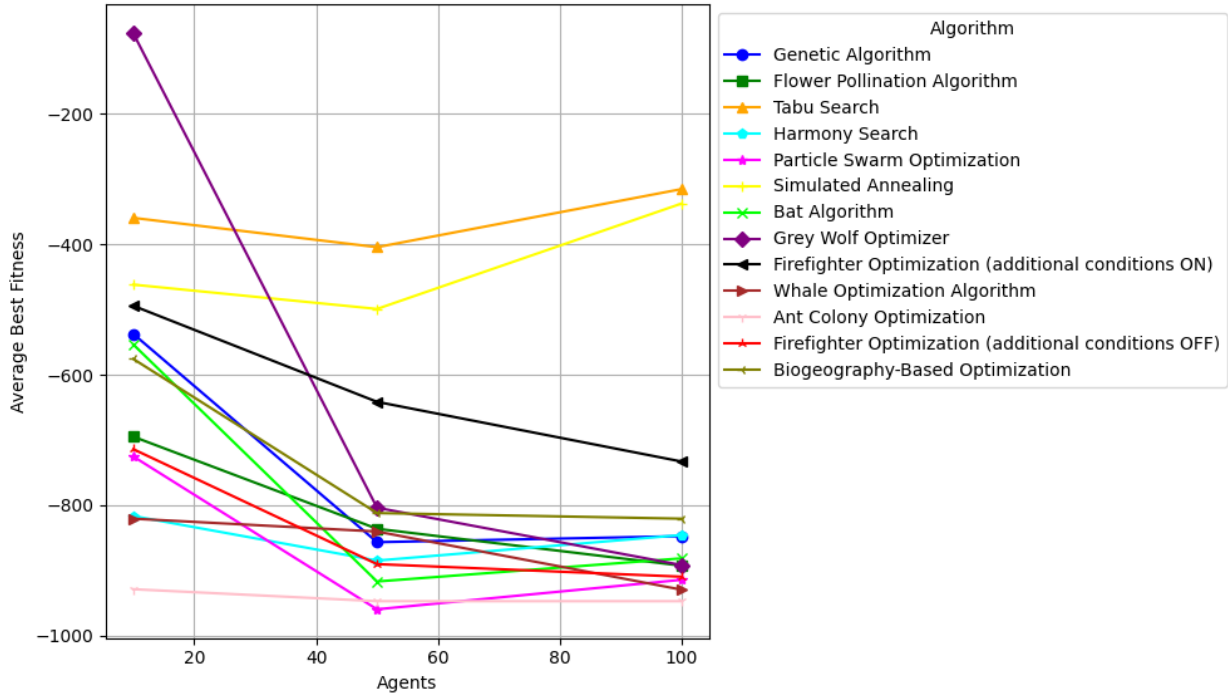

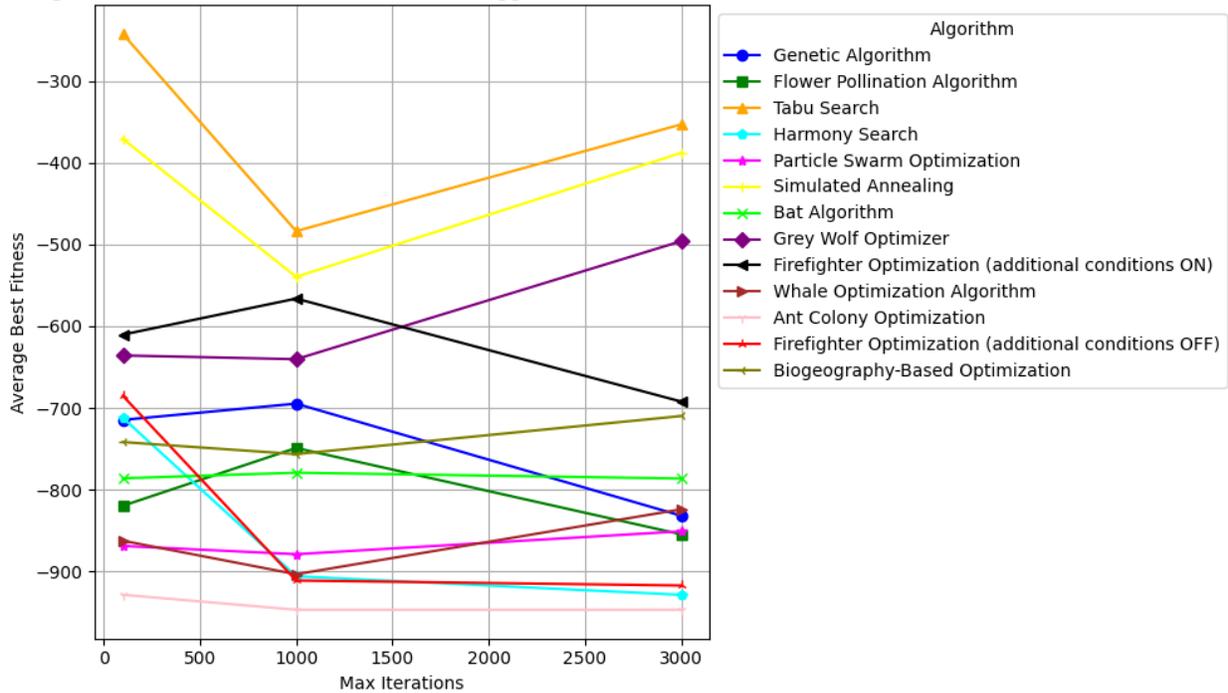

Fig. 7 Sample of trends across different experimental settings

We revisit the ranking performance of the optimization algorithms across the longest (and shortest) time to solve and that achieved the most (and least) accuracy. Similar to the case of 2D, the Firefly algorithm leads this category with an appearance frequency of 75, indicating it often requires the most time to solve the selected benchmark functions. On the other hand, the Ant Colony Optimization appears far less frequently with a count of 6, suggesting a quicker resolution time but possibly at the expense of other performance metrics like accuracy or search depth. Then, the Simulated Annealing dominates the fastest to solve category and hence demonstrates its ability to swiftly find solutions. In terms of accuracy, Cuckoo Search stands out with 22 appearances, followed by the FFO (additional conditions OFF) and Particle Swarm Optimization, each scoring 13 occurrences. The Ant Colony Optimization leads as the least accurate with 29 appearances, followed by the Bat Algorithm at 19, suggesting these algorithms might prioritize exploration or speed over precision.

Figure 8 paints a series of comparisons between selected algorithms. Figure 8a compares the average history fitness across all functions and algorithms. This graph shows the average fitness history across iterations for various optimization algorithms. The FFO (additional conditions OFF) starts with a rapid convergence compared to other algorithms, which is particularly noticeable against the backdrop of more gradual improvements shown by the Genetic Algorithm and Simulated Annealing. The FFO (additional conditions OFF) demonstrates a stabilization around 200 iterations, where its fitness value flatlines. This early convergence suggests that, on average, this algorithm is efficient in quickly finding a promising area of the search space.

While Fig. 8b compares the best fitness achieved in FFO, Tabu Search, the Bat, and the Grey Wolf Optimizer algorithms. In Fig. 8b, the performance of FFO (additional conditions OFF) is plotted across various scalable functions under different settings, showing stable performance for all functions, except the Rosenbrock, and especially under conditions of higher iterations and larger agent numbers. These spikes indicate that FFO (additional conditions OFF) can excel in complex, multimodal landscapes but show variable performance dependent on the function's characteristics and the search space complexity. Comparing this performance to other algorithms, such as Tabu Search, Bat, and Grey Wolf Optimizer, across the same benchmark functions shows generally stable performance as well, with some exceptions where fitness spikes, similar to the behavior seen in FFO. Figure 8c shows the performance of all algorithms in tackling continuous and non-continuous scalable benchmarking functions. The FFO (additional conditions OFF) shows competitive performance in all function types. These figures reinforce the performance of the newly proposed FFO algorithm against notable algorithms.

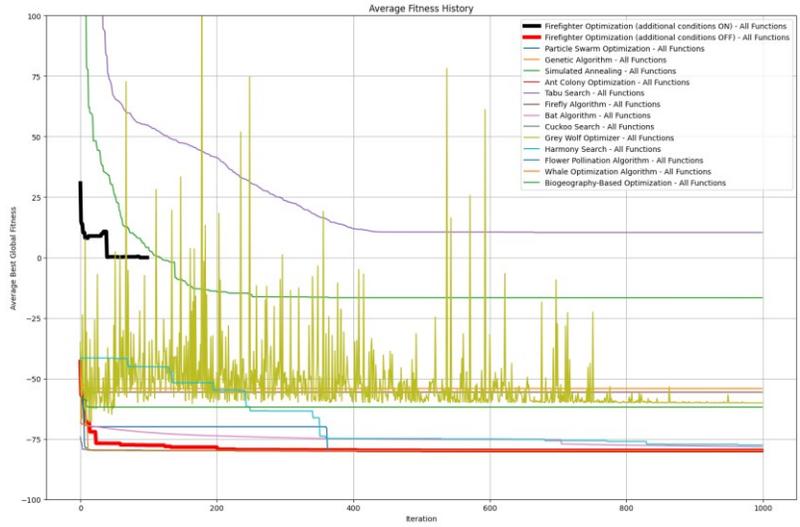

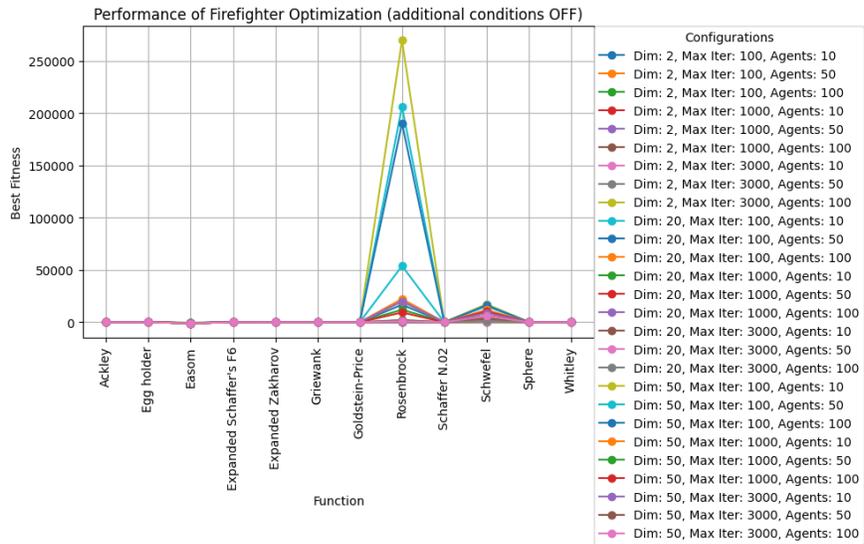

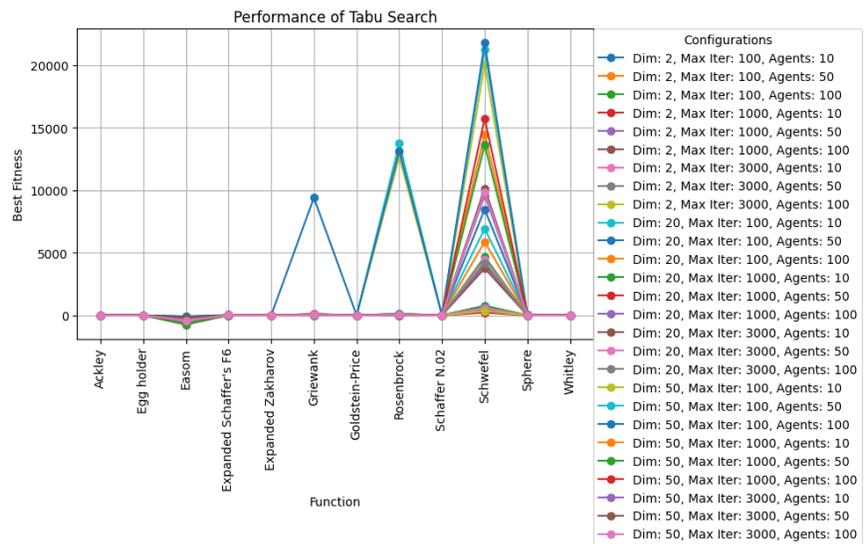

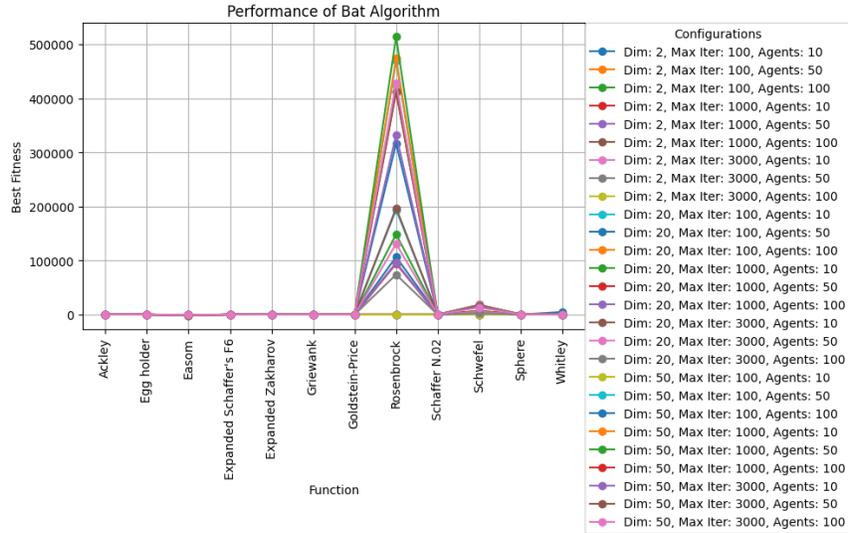

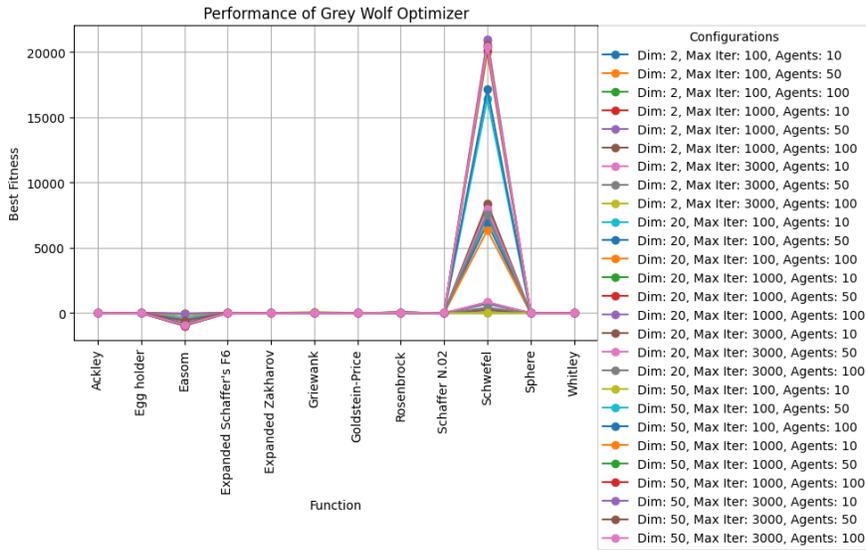

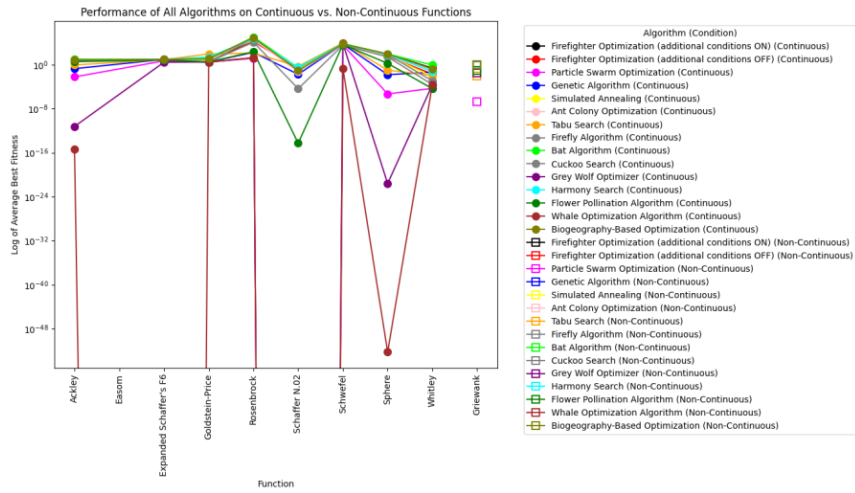

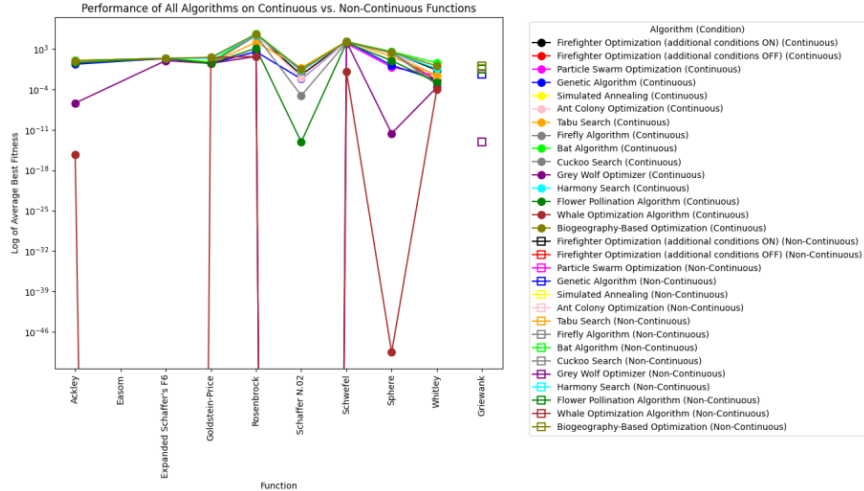

Fig. 8 Further cross examination between the FFO and other notable algorithms

The final set of results is depicted in radar plots and focuses on function attributes like scalability, separability, multimodality, differentiability, and performance in continuous function optimization. Each radar plot assesses these attributes quantitatively. This plot shows a sample that was produced by filtering results for specific settings (Dimension=2D, 20D, and 50D, Max Iterations=1000, Agents=100) and merging these results with predefined function attributes. The plots assess performance based on average best fitness, where each vertex on the radar represents an attribute. The plots then visually display how each algorithm performs across these attributes under the specific settings. These plots reveal distinct performance characteristics that generally show more modest peaks across the attributes, suggesting that some algorithms perform better in handling specific functions (i.e., Separable functions). For example, with higher dimensions, the FFO algorithm improves on the fronts of multimodality and separability.

These insights can also provide technical improvement areas for the selected algorithms (including the newly proposed algorithm). For example, this algorithm could focus on enhancing its performance in non-separable and multimodal landscapes, perhaps by integrating more adaptive search strategies or improving its handling of function differentiability through better derivative estimation or step size adjustment mechanisms. Additionally, efforts to boost scalability could be crucial, especially for handling higher-dimensional optimization problems more effectively.

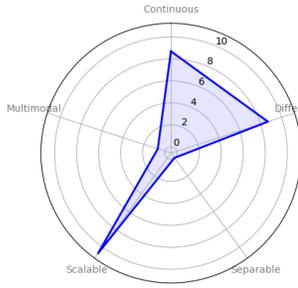
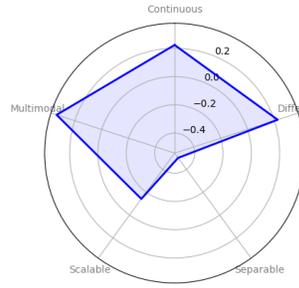
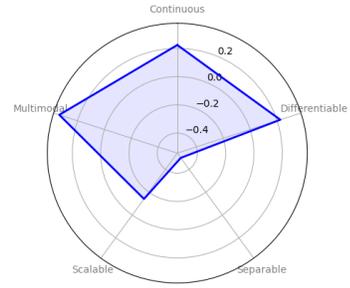
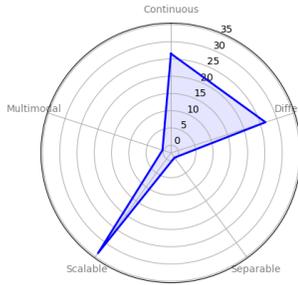
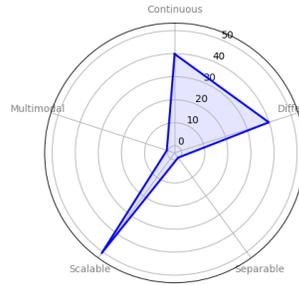
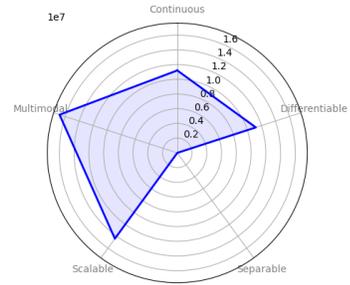
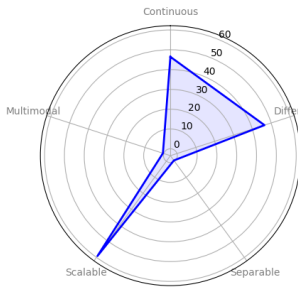
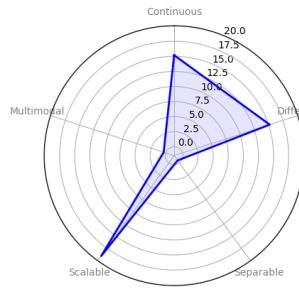
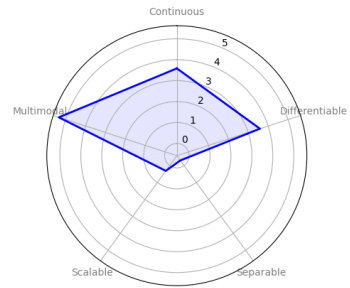
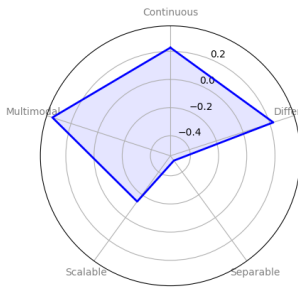
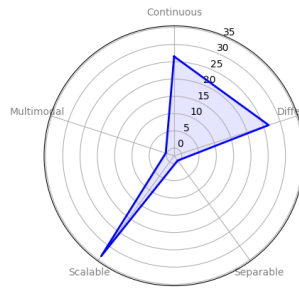
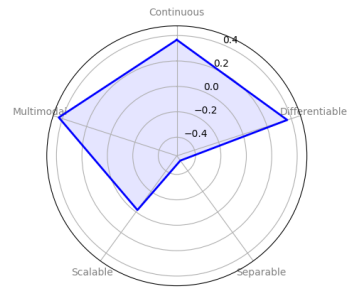
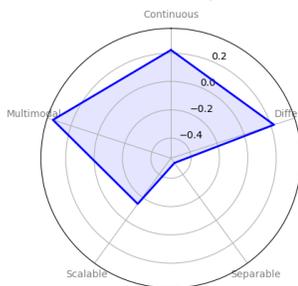
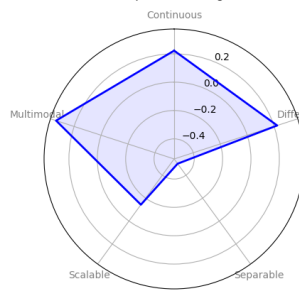
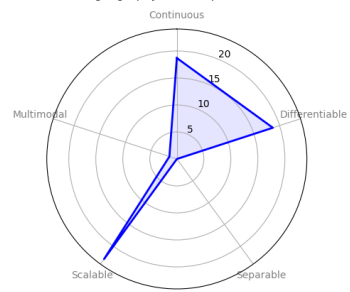

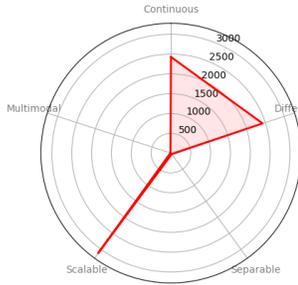
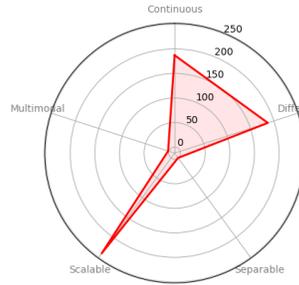
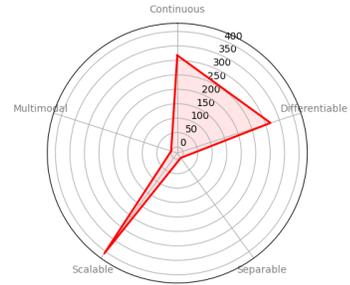
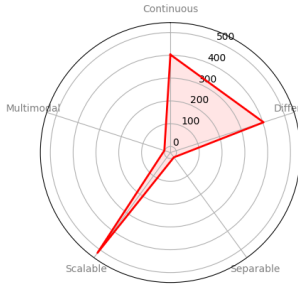
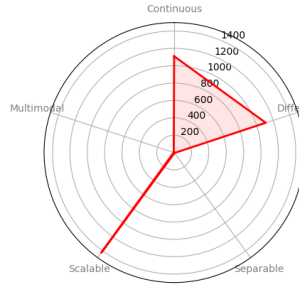
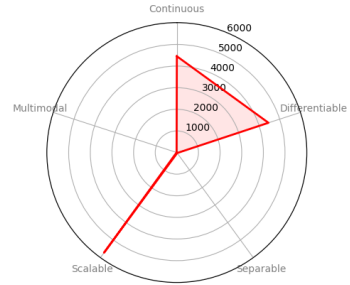
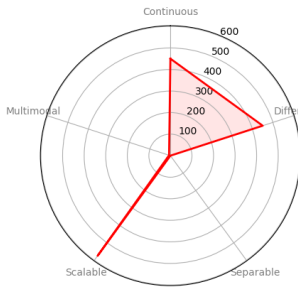
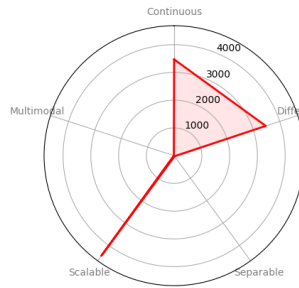
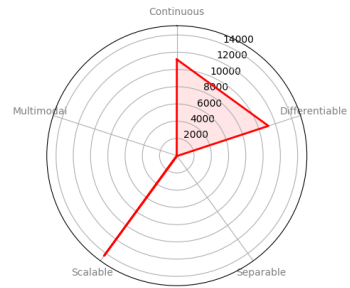
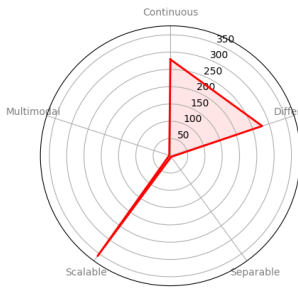
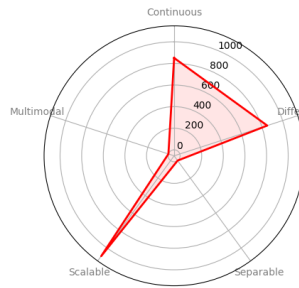
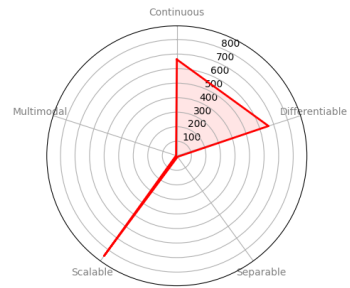
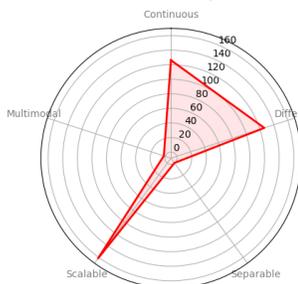
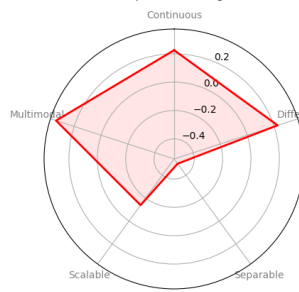
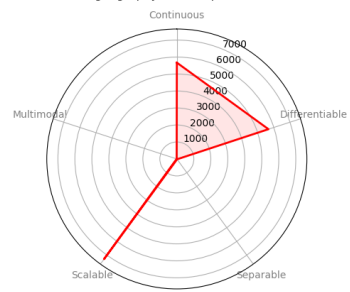

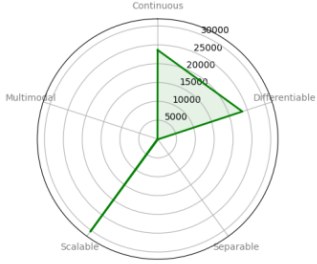
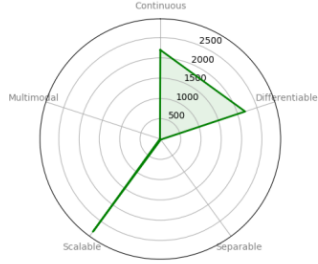
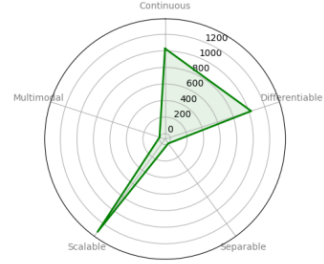
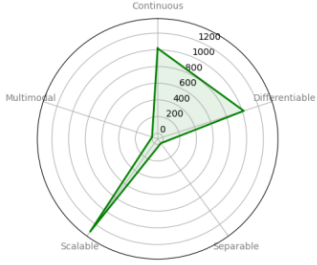
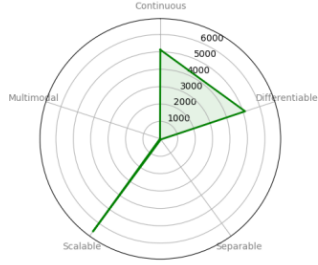
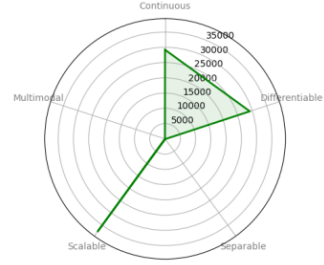
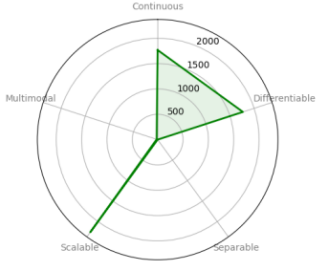
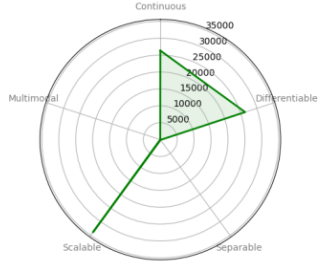
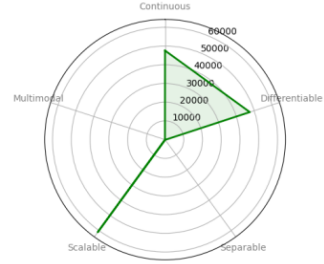
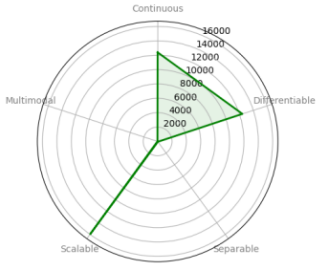
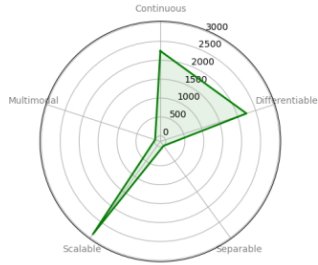
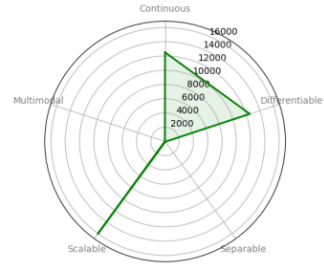
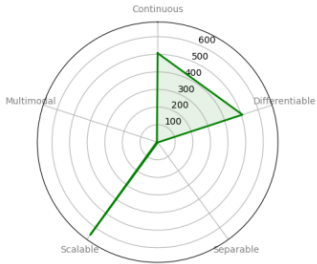
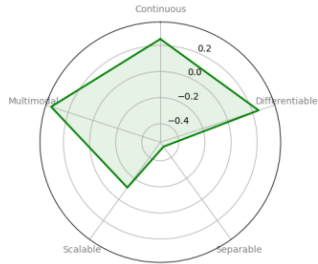
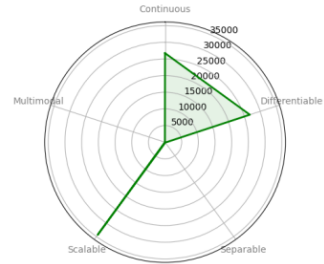

Fig. 9 Comparison of algorithms for function categories (Note: the settings are 100 agents and 1000 iterations for 2D, 20D, and 50D from top to bottom. Also, note that the scaling varies per algorithm for visibility purposes)

**6.0 Conclusions**

The Firefighter optimization (FFO) algorithm offers a hybrid metaheuristic approach for optimizing various problems. This algorithm was examined against 13 commonly used optimization algorithms, namely, the Ant Colony Optimization (ACO), Bat Algorithm (BA), Biogeography-Based Optimization (BBO), Flower Pollination Algorithm (FPA), Genetic Algorithm (GA), Grey Wolf Optimizer (GWO), Harmony Search (HS), Particle Swarm Optimization (PSO), Simulated Annealing (SA), Tabu Search (TS), and Whale Optimization Algorithm (WOA), and 24 benchmark functions. Our results demonstrate that FFO achieves comparative performance on multiple fronts (metrics). In addition, in 2D analysis, the FFO algorithm significantly improves best fitness and execution efficiency and large exploration capabilities. This algorithm ranked in the top 3 slots in terms of the best fitness and space covered. This algorithm also ranked first in the Distance per Unit Time metric. Similarly, the performance of the FFO algorithm is also notable in higher dimensions (20D and 50D) and maintains a top 5 performance.

**Data Availability**

Some or all data, models, or code that support the findings of this study are available from the corresponding author upon reasonable request.

Firefighter optimization (FFO) can be accessed from [**to be added**].

**Conflict of Interest**

The authors declare no conflict of interest.

and Open Challenges. (arXiv:2007.00541v1 [cs.NE]), ArXiv Comput. Sci. (2020).

[8] K.-L. Du, M.N.S. Swamy, Search and Optimization by Metaheuristics, 2016. https://doi.org/10.1007/978-3-319-41192-7.

[9] M. Dorigo, M. Birattari, T. Stützle, Ant colony optimization artificial ants as a computational intelligence technique, IEEE Comput. Intell. Mag. (2006). https://doi.org/10.1109/CI-M.2006.248054.

[10] J.E. Bell, P.R. McMullen, Ant colony optimization techniques for the vehicle routing problem, Adv. Eng. Informatics. (2004). https://doi.org/10.1016/j.aei.2004.07.001.

[11] X.S. Yang, A.H. Gandomi, Bat algorithm: A novel approach for global engineering optimization, Eng. Comput. (Swansea, Wales). (2012). https://doi.org/10.1108/02644401211235834.

[12] P.W. Tsai, J.S. Pan, B.Y. Liao, M.J. Tsai, V. Istanda, Bat algorithm inspired algorithm for solving numerical optimization problems, in: Appl. Mech. Mater., 2012. https://doi.org/10.4028/www.scientific.net/AMM.148-149.134.

[13] D. Simon, Biogeography-based optimization, IEEE Trans. Evol. Comput. (2008). https://doi.org/10.1109/TEVC.2008.919004.

[14] R.A. Gupta, R. Kumar, A.K. Bansal, BBO-based small autonomous hybrid power system optimization incorporating wind speed and solar radiation forecasting, Renew. Sustain. Energy Rev. (2015). https://doi.org/10.1016/j.rser.2014.09.017.

[15] X.S. Yang, S. Deb, Cuckoo search via Lévy flights, in: 2009 World Congr. Nat. Biol. Inspired Comput. NABIC 2009 - Proc., 2009. https://doi.org/10.1109/NABIC.2009.5393690.

[16] A.H. Gandomi, X.S. Yang, A.H. Alavi, Cuckoo search algorithm: A metaheuristic approach to solve structural optimization problems, Eng. Comput. (2013). https://doi.org/10.1007/s00366-011-0241-y.

[17] X.-S. Yang, A. Slowik, Firefly Algorithm, Swarm Intell. Algorithms. (2020) 163–174. https://doi.org/10.1201/9780429422614-13.

[18] H. Xie, L. Zhang, C.P. Lim, Y. Yu, C. Liu, H. Liu, J. Walters, Improving K-means clustering with enhanced Firefly Algorithms, Appl. Soft Comput. J. (2019). https://doi.org/10.1016/j.asoc.2019.105763.

[19] X.S. Yang, M. Karamanoglu, X. He, Flower pollination algorithm: A novel approach for multiobjective optimization, Eng. Optim. (2014). https://doi.org/10.1080/0305215X.2013.832237.

[20] S. Lalljith, I. Fleming, U. Pillay, K. Naicker, Z.J. Naidoo, A.K. Saha, Applications of Flower Pollination Algorithm in Electrical Power Systems: A Review, IEEE Access. (2022). https://doi.org/10.1109/ACCESS.2021.3138518.

[21] J.H. Holland, Adaptation in natural and artificial systems : an introductory analysis with applications to biology, control, and artificial intelligence, 1975.